\documentclass[11pt]{article}
\usepackage[truedimen,margin=25truemm]{geometry}
\usepackage{microtype}
\usepackage{graphics}

\usepackage{times}
\usepackage{xcolor}
\usepackage{amsmath}
\usepackage{amsfonts}
\usepackage{amssymb}
\usepackage{url}
\usepackage{multirow}
\usepackage{mathtools}
\usepackage{booktabs}
\usepackage{graphicx}
\usepackage{epstopdf}
\usepackage[ruled]{algorithm2e}
\usepackage{algorithmic}
\usepackage[numbers]{natbib}

\newcommand{\refalg}[1]{Algorithm \ref{#1}}
\newcommand{\refeqn}[1]{ \eqref{#1}}
\newcommand{\reffig}[1]{Figure \ref{#1}}
\newcommand{\reftbl}[1]{Table \ref{#1}}
\newcommand{\refsec}[1]{Section \ref{#1}}

\newcommand{\system}{SIITA}

\newcommand{\systemfull}{Side Information infused Incremental Tensor Analysis}

\newcommand\norm[1]{\left\lVert#1\right\rVert}

\DeclareMathAlphabet\ten{OMS}{cmsy}{b}{n} 

\def\mat#1{\mbox{\bf #1}}

\def\set#1{\mathcal{#1}}

\title{Inductive Framework for Multi-Aspect Streaming\\ Tensor Completion with Side Information}

\author{
Madhav Nimishakavi$^1$, 
Bamdev Mishra$^2$, 
Manish Gupta$^2$, 
Partha Talukdar$^1$
\\ 
$^1$ Indian Institute of Science, India \\
$^2$ Microsoft, India  \\
madhav@iisc.ac.in,
bamdevm@microsoft.com,
gmanish@microsoft.com,
ppt@iisc.ac.in
}

\date{ }
\begin{document}
\maketitle
\begin{abstract}
Low rank tensor completion is a well studied problem and has applications in various fields. However, in many real world applications the data is dynamic, i.e., new data arrives at different time intervals. As a result, the tensors used to represent the data grow in size. Besides the tensors, in many 
real world scenarios, side information is also available in the form of matrices which also grow in size with time. The problem of predicting missing values in the dynamically growing tensor is called dynamic tensor completion. Most of the previous work in dynamic tensor completion make an assumption that the tensor grows only in one mode. To the best of our Knowledge, there is no previous work which incorporates side information with dynamic tensor completion. 
We bridge this gap in this paper by proposing a dynamic tensor completion framework called \systemfull{} (\system{}), which incorporates side information and works for general incremental tensors. We also show how non-negative constraints can be incorporated with \system{}, which is essential for mining interpretable latent clusters.
We carry out extensive experiments on multiple real world datasets to demonstrate the effectiveness of \system{} in various different settings. 
\end{abstract}

\section{Introduction}
\label{sec:intro}
Low rank tensor completion is a well-studied problem and has various applications in the fields of recommendation systems \cite{Symeonidis2008}, link-prediction \cite{Ermis2015}, compressed sensing \cite{Cichocki2015}, to name a few. Majority of the previous works focus on solving the problem in a static setting \cite{Filipovic2015,Guo2017,Kasai2016a}. However, most of the real world data is dynamic, for example in an online  movie recommendation system the number of users and movies increase with time. It is prohibitively expensive to use the static algorithms for dynamic data. Therefore, there has been an increasing interest in developing algorithms for dynamic low-rank tensor completion \cite{Kasai2016,Mardani2015,Song2017}.

Usually in many real world scenarios, besides the tensor data,  additional side information is also available, e.g., in the form of matrices. In the dynamic scenarios, the side information grows with time as well. For instance, movie-genre information in the movie recommendation etc. There has been considerable amount of work in incorporating side information into tensor completion \cite{Narita2011,Ge2016}. However, the previous works on incorporating side information deal with the static setting. In this paper, we propose a dynamic low-rank tensor completion model that incorporates side information growing with time. 

\begin{figure}[t]
\centering
\begin{tabular}{c}
\noindent\begin{minipage}[b]{0.5\hsize}
\centering
\includegraphics[width=1\textwidth]{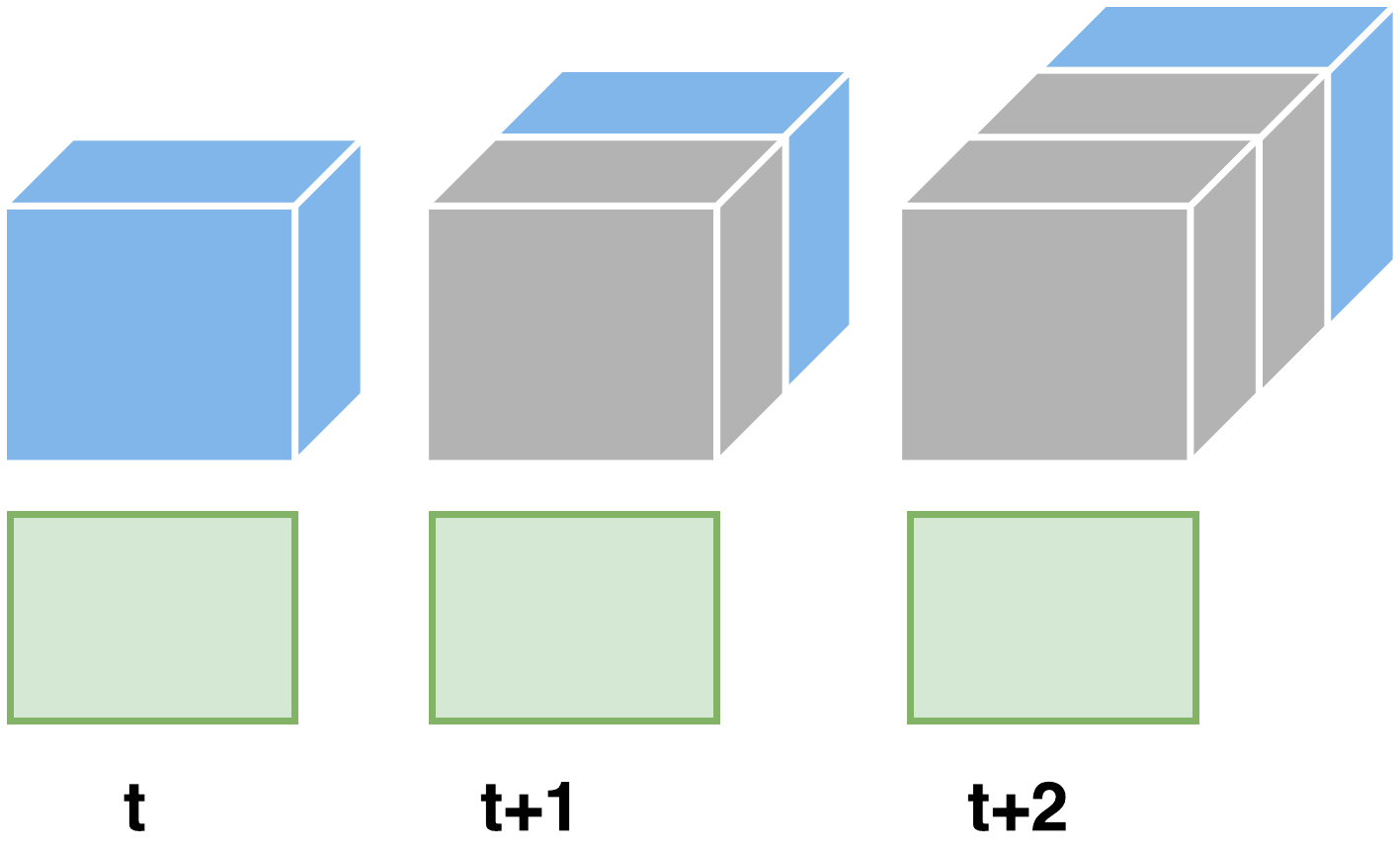}\\
{\small(a) Streaming tensor sequence with side information.}
\end{minipage}
 \begin{minipage}[b]{0.5\hsize}
\centering
\includegraphics[width=1\textwidth]{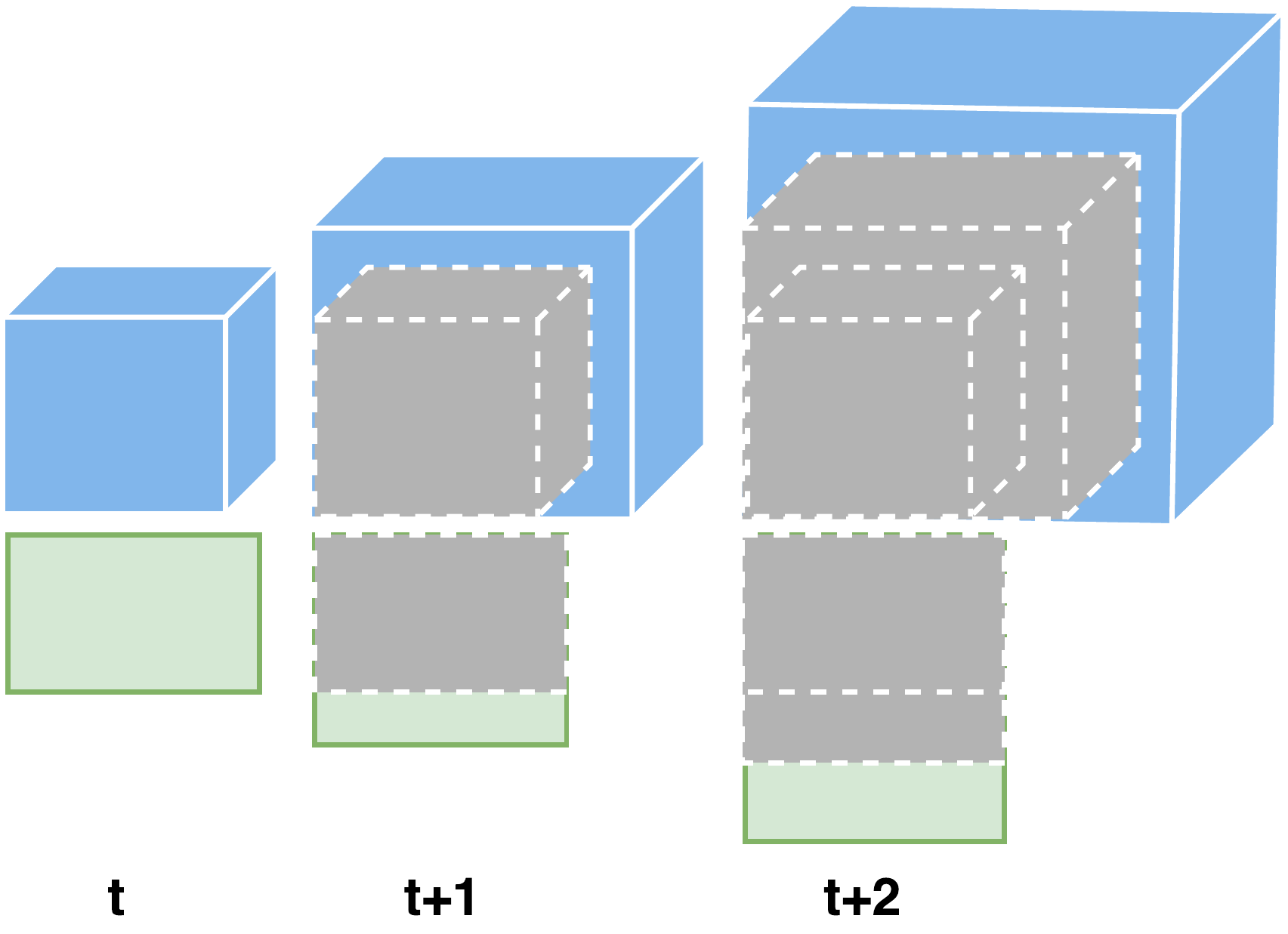}\\
{\small (b) Multi-aspect streaming tensor sequence with side information.}
\end{minipage}
\end{tabular}
\caption{Illustration of streaming and multi-aspect streaming sequences with side information. The blue block represents the tensor at time step and the green block represents the side information. The blocks in grey represent the data at previous time steps.
For easy understanding, we show side information along only one mode.}
\label{fig:illustartion}
\end{figure}

Most of the current dynamic tensor completion algorithms work in the streaming scenario, i.e., the case where the tensor grows only in one mode, which is usually the time mode. In this case, the side information is a static matrix. Multi-aspect streaming scenario \cite{Fanaee-T2015,Song2017}, on the other hand, is a more general framework, where the tensor grows in all the modes of the tensor. In this setting, the side information matrices also grow. \reffig{fig:illustartion} illustrates the difference between streaming and multi-aspect streaming scenarios with side information.

Besides side information, incorporating nonnegative constraints into tensor decomposition is desirable in an unsupervised setting. Nonnegativity is essential for discovering interpretable clusters \cite{Hyvoenen2008, Murphy2012}. Nonnegative tensor learning is explored for applications in computer vision \cite{Shashua2005,Kim2007}, unsupervised induction of relation schemas \cite{Nimishakavi2016}, to name a few.
Several algorithms for online Nonnegative Matrix Factorization (NMF) exist in the literature \cite{Lefevre2011, Zhao2017}, but algorithms for nonnegative online tensor decomposition with side information are not explored to the best of our knowledge. We also fill this gap by showing how nonnegative constraints can be enforced on the decomposition learned by our proposed framework \system{}.

In this paper, we work with the more general multi-aspect streaming scenario and make the following contributions:
\begin{itemize}
\item  Formally define the problem of multi-aspect streaming tensor completion with side information.
\item  Propose a Tucker based framework \systemfull (\system{}) for the problem of multi-aspect streaming tensor completion with side information. We employ a stochastic gradient descent (SGD) based algorithm for solving the optimization problem.
\item Incorporate nonnegative constraints with \system{} for discovering the underlying clusters in unsupervised setting.
\item  Demonstrate the effectiveness of \system{} using extensive experimental analysis on multiple real-world datasets in all the settings.
\end{itemize}

The organization of the paper is as follows. In Section \ref{sec:prelim}, we introduce the definition of multi-aspect streaming tensor sequence with side information and discuss our proposed framework \system{} in Section \ref{sec:mast_si}. We also discuss how nonnegative constraints can be incorporated into \system{} in Section \ref{sec:mast_si}.  The experiments are shown in Section \ref{sec:exp}, where \system{} performs effectively in various settings. All our codes are implemented in Matlab, and can be found at \texttt{\url{https://madhavcsa.github.io/}}.

\section{Related Work}
\label{sec:rel}

\begin{table*}[bt]
 \centering
 \small
 \caption{ Summary of different tensor streaming algorithms.\label{tbl:bsline_table}}
 \begin{tabular}{p{3.5cm} p{2cm} p{2cm} p{2cm} p{2cm} p{2cm} p{3.2cm}}
  \toprule
  Property & TeCPSGD\cite{Mardani2015} &OLSTEC \cite{Kasai2016} & MAST \cite{Song2017}  & AirCP \cite{Ge2016} & \system{} (this paper) \\
  \midrule 
  Streaming 		 &	\multicolumn{1}{c}{\checkmark}& 	\multicolumn{1}{c}{\checkmark}	&\multicolumn{1}{c}{\checkmark}	&				&\multicolumn{1}{c}{\checkmark}  \\
  Multi-Aspect Streaming &	&		&	\multicolumn{1}{c}{\checkmark}	&				&\multicolumn{1}{c}{\checkmark}  \\
  Side Information	&	&		&		&	\multicolumn{1}{c}{\checkmark}			& \multicolumn{1}{c}{\checkmark}	 \\
  Sparse Solution	&	&		&		&				&\multicolumn{1}{c}{\checkmark}	  \\
  \bottomrule
 \end{tabular} 
\end{table*}


{\bf Dynamic Tensor Completion :}  \cite{Sun2006,Sun2008} introduce the concept of dynamic tensor analysis by proposing multiple Higher order SVD based algorithms, namely Dynamic Tensor Analysis (DTA), Streaming Tensor Analysis (STA) and Window-based Tensor Analysis (WTA) for the streaming scenario.
 \cite{Nion2009} propose two adaptive  online algorithms for CP decomposition of $3$-order tensors.  \cite{Yu2015}  propose an accelerated online algorithm for tucker factorization in streaming scenario, while an accelerated online algorithm for CP decomposition is developed in \cite{Zhou2016}.

A significant amount of research work is carried out for dynamic tensor decompositions, but work focusing on the problem of dynamic tensor completion is relatively less explored. Work by  \cite{Mardani2015} can be considered a pioneering work in dynamic tensor completion. They propose a streaming tensor completion algorithm based on CP decomposition.
Recent work by  \cite{Kasai2016} is an accelerated second order Stochastic Gradient Descent (SGD) algorithm for streaming tensor completion based on CP decomposition.
 \cite{Fanaee-T2015} introduces the problem of multi-aspect streaming tensor analysis by proposing a histogram based algorithm. Recent work by \cite{Song2017} is a more
general framework for multi-aspect streaming tensor completion. 

{\bf Tensor Completion with Auxiliary Information :}  \cite{Acar2011} propose a Coupled Matrix Tensor Factorization (CMTF) approach for incorporating additional side information, similar ideas are also explored in
\cite{Beutel2014} for factorization on hadoop and in \cite{Ermis2015a} for link prediction in heterogeneous data. \cite{Narita2011} propose with-in mode and cross-mode regularization methods for incorporating similarity side information matrices into factorization. Based on similar ideas, \cite{Ge2016} propose AirCP, a CP-based tensor completion algorithm. 

\cite{Welling2001} propose nonnegative tensor decmpositon by incorporating nonnegative constraints into CP decomposition. Nonnegative CP decomposition is explored for applications in computer vision in \cite{Shashua2005}. Algorithms for nonnegative Tucker decomposition are proposed in \cite{Kim2007} and for sparse nonnegative Tucker decomposition are proposed in \cite{Morup2008}. However, to the best our knowledge, nonnegative tensor decomposition algorithms do not exist for dynamic settings, a gap we fill in this paper.

Inductive framework for matrix completion with side information is proposed in \cite{jain2013,Natarajan2014,Si2016}, which has not been explored for tensor completion to the best of our knowledge. In this paper, we propose an online inductive framework for multi-aspect streaming tensor completion.

\reftbl{tbl:bsline_table} provides details about the differences between our proposed \system{} and various baseline tensor completion algorithms.

\section{Preliminaries}
\label{sec:prelim}


An $N^{th}$-order or $N$-mode tensor is an $N$-way array. We use boldface calligraphic letters to represent tensors (e.g., $\ten{X}$), boldface uppercase to represent matrices (e.g., $\mat{U}$), and boldface lowercase to represent vectors (e.g., $\mat{v}$). $\ten{X}[i_1, \cdots, i_N]$ represents the entry of $\ten{X}$ indexed by $[i_1, \cdots, i_N]$. 
~\\
{\bf Definition 1 (Coupled Tensor and Matrix) \cite{Song2017}}: A matrix  and a tensor are called coupled if they share a mode. For example, a ${user} \times {movie} \times { time}$ tensor and a ${movie} \times {genre}$ matrix are coupled along the {movie} mode.
~\\
{\bf Definition 2 (Tensor Sequence) \cite{Song2017}}: 
A sequence of  $N^{th}$-order  tensors $\ten{X}^{(1)}, \ldots , \ten{X}^{(t)}, \dots$ is called a tensor sequence denoted as $\{\ten{X}^{(t)}\}$, where each $\ten{X}^{(t)} \in \mathbb{R}^{I_1^t \times I_2^t \times \ldots \times I_N^t}$ at time instance $t$. 
~\\
{\bf Definition 3 (Multi-aspect streaming Tensor Sequence) \label{def:mast-seq}\cite{Song2017}}: 
A tensor sequence of $N^{th}$-order tensors $\{\ten{X}^{(t)}\}$ is called a multi-aspect streaming tensor sequence  if for any $t \in \mathbb{Z}^{+}$, $\ten{X}^{(t-1)} \in \mathbb{R}^{I_1^{t-1} \times I_2^{t-1} \times \ldots \times I_N^{t-1}}$ is the sub-tensor of $\ten{X}^{(t)} \in \mathbb{R}^{I_1^t \times I_2^t \times \ldots \times I_N^t}$, i.e., 
	\[
		\ten{X}^{(t-1)} \subseteq \ten{X}^{(t)},~\mathrm{where}~I_{i}^{t-1} \le I_{i}^{t},~\forall 1 \le i \le N.
	\]
Here, $t$ increases with time, and $\ten{X}^{(t)}$  is the snapshot tensor of this sequence at time $t$.
~\\
{\bf Definition 4 (Multi-aspect streaming Tensor Sequence with Side Information) }: Given a time instance $t$, let $\mat{A}_{i}^{(t)} \in \mathbb{R}^{I_{i}^{t} \times M_i}$ be a side information (SI) matrix corresponding to the $i^{th}$ mode of $\ten{X}^{(t)}$ (i.e., rows of $\mat{A}_{i}^{(t)}$ are coupled along $i^{th}$ mode of $\ten{X}^{(t)}$). While the number of rows in the SI matrices along a particular mode $i$ may increase over time, the number of columns remain the same, i.e., $M_i$ is not dependent on time. In particular, we have,
	\begin{align*}
   		\mat{A}_{i}^{(t)}  &= 
			   \begin{bmatrix}
				    \mat{A}_{i}^{(t-1)} \\
				    \Delta_{i}^{(t)}
			   \end{bmatrix},~\mathrm{where}~\Delta_{i}^{(t)} \in \mathbb{R}^{[I_i^{(t)} - I_{i}^{(t-1)}] \times M_i}.
	 \end{align*} 
Putting side information matrices of all the modes together,  we get the side information set $\set{A}^{(t)}$,
	\[
		\set{A}^{(t)} = \{\mat{A}_{1}^{(t)}, \ldots, \mat{A}_{N}^{(t)}\}.
	\]
Given an $N^{th}$-order multi-aspect streaming tensor sequence $\{\ten{X}^{(t)}\}$, we define a  multi-aspect streaming tensor sequence with side information as $\{(\ten{X}^{(t)}, \set{A}^{(t)})\}$.

We note that all modes may not have side information available. In such cases, an identity matrix of appropriate size may be used as $\mat{A}_{i}^{(t)}$, i.e., $\mat{A}_{i}^{(t)} = \mat{I}^{I_{i}^{t} \times I_{i}^{t}}$, where $M_i = I_{i}^{t}$.

The problem of multi-aspect streaming tensor completion with side information is formally defined as follows:

\begin{center}
\framebox{\parbox{\dimexpr\linewidth-2\fboxsep-2\fboxrule}{  
{\bf Problem Definition}:  Given a multi-aspect streaming tensor sequence with side information  $\{(\ten{X}^{(t)},  \set{A}^{(t)})\}$, the goal is to predict the missing values in $\ten{X}^{(t)}$ by utilizing  only entries in the relative complement $\ten{X}^{(t)} \setminus \ten{X}^{(t-1)}$ and the available side information $\set{A}^{(t)}$. 
}}
\end{center}


\section{Proposed Framework \system{}}   
\label{sec:mast_si}

%
%

In this section, we discuss the proposed framework \system{} for the problem of multi-aspect streaming tensor completion with side information. 
Let $\{(\ten{X}^{(t)}, \mathcal{A}^{(t)}) \}$ be an $N^{th}$-order multi-aspect streaming tensor sequence with side information. Assuming that, at every time step, $\ten{X}^{(t)}[i_1, i_2, \cdots, i_N]$ are only observed for some indices $[i_1, i_2, \cdots, i_N] \in \Omega$, where $\Omega$ is a subset of the complete set of indices $[i_1, i_2, \cdots, i_N]$.
Let the sparsity operator $\mathcal{P}_{\Omega} $ be defined as:
\begin{equation*}
 \mathcal{P}_{\Omega}[i_1, i_2, \cdots, i_N] = 
 \begin{cases}
  \ten{X}[i_1, \cdots, i_N], & \text{if}\ [i_1, \cdots, i_N] \in \Omega\\
  0, & \text{otherwise}.
 \end{cases}
\end{equation*}


Tucker tensor decomposition \cite{Kolda2009}, is a form of higher-order PCA for tensors. It decomposes an $N^{th}$-order tensor $\ten{X}$ into a core tensor multiplied by a matrix along each mode as follows
\begin{equation*}
    \ten{X} \approx \ten{G} \times_1 \mat{U}_1 \times_2 \mat{U}_2 \times_3 \cdots \mat{U}_N,
\end{equation*}
where, $\mat{U}_i \in \mathbb{R}^{I_i \times r_i}, i=1:N$ are the factor matrices and can be thought of as principal components in each mode. The tensor $\ten{G} \in \mathbb{R}^{r_1 \times r_2 \times \cdots r_N}$ is called the \emph{core tensor}, which shows the interaction between different components. $(r_1, r_2, \cdots, r_N)$ is the (multilinear) rank of the tensor. The $i$-mode matrix product of a tensor  $\ten{X} \in \mathbb{R}^{I_1 \times I_2 \times \cdots I_N}$ with a matrix $\mat{P} \in \mathbb{R}^{r \times I_i}$ is denoted by $\ten{X}\times_i \mat{P}$, more details can be found in \cite{Kolda2009}. {The standard approach of incorporating side information while learning factor matrices in Tucker decomposition is by using an additive term as a regularizer \cite{Narita2011}. However, in an online setting the additive side information term poses challenges as the side information matrices are also dynamic.} Therefore, we propose the following fixed-rank {\it inductive framework} for recovering missing values in $\ten{X}^{(t)}$, at every time step $t$:
\begin{equation}
\label{eqn:ind_tucker}
 \underset{\mat{U}_i \in \mathbb{R}^{M_i \times r_i},  i = 1:N}{\min_{\ten{G} \in \mathbb{R}^{r_1 \times \ldots \times r_N}}}  F(\ten{X}^{(t)}, \mathcal{A}^{(t)}, \ten{G}, \{\mat{U}_i\}_{i=1:N}),
\end{equation}
where
\begin{multline}
\label{eqn:F_U_G}
  F(\ten{X}^{(t)}, \mathcal{A}^{(t)}, \ten{G}, \{\mat{U}_n\}_{i=1:N}) = 
  \norm{\mathcal{P}_{\Omega}(\ten{X}^{(t)}) -
 \mathcal{P}_{\Omega}(\ten{G} \times_1 \mat{A}_1^{(t)}\mat{U}_1 \times_2 \ldots \times_N \mat{A}_N^{(t)}\mat{U}_N)}_F^2 \\
  + \lambda_g \norm{\ten{G}}_F^2 + \sum_{i=1}^{N}\lambda_i \norm{\mat{U}_i}_F^2.
\end{multline}
$\norm{\cdot}_F$ is the Frobenius norm, $\lambda_g > 0$ and $\lambda_i > 0, i=1:N$ are the regularization weights. Conceptually, the inductive framework models the ratings of the tensor as a weighted scalar product of the side information matrices. Note that (\ref{eqn:ind_tucker}) is a generalization of the inductive matrix completion framework \cite{jain2013,Natarajan2014,Si2016}, which has been effective in many applications.

The inductive tensor framework has two-fold benefits over the typical approach of incorporating side information as an additive term. The use of $\mat{A}_i \mat{U}_i$ terms in the factorization reduces the dimensionality of variables from $\mat{U}_i \in \mathbb{R}^{I_i \times r_i}$ to $\mat{U}_i \in \mathbb{R}^{M_i \times r_i}$ and typically $M_i \ll I_i$. As a result, computational time required for computing the gradients and updating the variables decreases remarkably. Similar to \cite{Kim2007}, we define 
\begin{equation*}
 \mat{U}_i^{(\backslash n)} = \big[ \mat{A}_{i-1}^{(t)} \mat{U}_{i-1} \otimes \ldots \otimes \mat{A}_{1}^{(t)} \mat{U}_{1} \otimes \ldots \otimes 
 \mat{A}_{N}^{(t)} \mat{U}_{N} \otimes \ldots \otimes \mat{A}_{i+1}^{(t)} \mat{U}_{i+1} \big ] \nonumber ,
\end{equation*}
which collects Kronecker products of mode matrices except for $\mat{A}_i\mat{U}_i$ in a backward cyclic manner.

The gradients for \eqref{eqn:ind_tucker} wrt $\mat{U}_i$ for $i = 1:N$ and $\ten{G}$ can be computed as following:
\begin{equation}\label{eqn:gradu}
\begin{array}{lll}
\displaystyle \frac{\partial F}{\partial \mat{U}_i} = -(\mat{A}_i^{(t)})^\top \ten{R}_{(i)}^{(t)} \mat{U}_i^{(\backslash n)} \ten{G}_{(i)}^\top + 2\lambda_i \mat{U}_i \\
\\
 \displaystyle \frac{\partial F}{\partial \ten{G}} = - \ten{R}^{(t)} \times_1 (\mat{A}_1^{(t)}\mat{U}_1)^{\top} \times_2 \ldots \times_N  (\mat{A}_N^{(t)}\mat{U}_N)^{\top}+2\lambda_g \ten{G}, 
 \end{array}
\end{equation}
where
\begin{equation*}
\label{eqn:res}
\ten{R}^{(t)} = \ten{X}^{(t)} - \ten{G} \times_1 \mat{A}_1^{(t)} \mat{U}_1 \times_2 \ldots \times_N \mat{A}_N^{(t)} \mat{U}_N .
\end{equation*}

%

By updating the variables using gradients given in (\ref{eqn:gradu}), we can recover the missing entries in $\ten{X}^{(t)}$ at every time step $t$, however that is equivalent to performing a static tensor completion  at every time step. Therefore, we need an incremental scheme for updating the variables. Let $\mat{U}_i^{(t)}$ and $\ten{G}^{(t)}$ represent the variables at time step $t$, then
\begin{equation}
\begin{array}{ll}
 F(\ten{X}^{(t)}, \mathcal{A}^{(t)}, \ten{G}^{(t-1)}, \{\mat{U}_i^{(t-1)}\}_{i=1:N}) =\\
 \qquad F(\ten{X}^{(t-1)}, \mathcal{A}^{(t-1)}, \ten{G}^{(t-1)}, \{\mat{U}_i^{(t-1)}\}_{i=1:N})\ \  +  \\ 
\qquad  F(\ten{X}^{(\Delta t)}, \mathcal{A}^{(\Delta t)}, \ten{G}^{(t-1)}, \{\mat{U}_i^{(t-1)}\}_{i=1:N}), 
 \end{array}
\end{equation}
since $\ten{X}^{(t-1)}$ is recovered at the time step $t$-$1$, the problem is equivalent to using only 
\[
F^{(\Delta t)} = F(\ten{X}^{(\Delta t)}, \mathcal{A}^{(\Delta t)}, \ten{G}^{(t-1)}, \{\mat{U}_i^{(t-1)}\}_{i=1:N}), 
\]
for updating the variables at time step $t$. 

We propose to use the following approach to update the variables at every time step $t$, i.e.,
\begin{equation}\label{eqn:update_u}
\begin{array}{lll}
 \mat{U}_i^{(t)} = \mat{U}_i^{(t-1)} - \gamma \frac{\partial F^{(\Delta t)}}{\partial \mat{U}_i^{(t-1)}}, i = 1:N \\
  \ten{G}^{(t)} =  \ten{G}^{(t-1)} - \gamma \frac{\partial F^{(\Delta t)}}{\partial \ten{G}^{(t-1)}},
\end{array}
\end{equation}
where $\gamma$ is the step size for the gradients. $\ten{R}^{(\Delta t)}$, needed for computing the gradients of $F^{(\Delta t)}$, is given by 
\begin{equation}\label{eqn:r_delta}
\ten{R}^{(\Delta t)}  = \ten{X}^{(\Delta t)} - \ten{G}^{(t-1)} \times_1 \mat{A}_1^{(\Delta t)}\mat{U}_1^{(t-1)} \times_2 \ldots \times_N \mat{A}_N^{(\Delta t)}\mat{U}_N^{(t-1)}.
\end{equation}


\begin{algorithm}[t]
\small
\caption{Proposed \system{} Algorithm}\label{alg:simast}
 \SetKwInOut{Input}{Input}
 \SetKwInOut{Output}{Return}
 
 \Input{$ \{\ten{X}^{(t)}, \mathcal{A}^{(t)}\}, \lambda_i, i = 1:N ,(r_1, \ldots, r_N) $}
 Randomly initialize $\mat{U}_i^{(0)} \in \mathbb{R}^{M_i \times r_i}, i = 1:N$ and $\ten{G}^{(0)} \in \mathbb{R}^{r_i \times \ldots \times r_N}$ ;\\
 \For{t = 1, 2, \ldots}
 {
 $\mat{U}_i^{(t)_{0}} \coloneqq \mat{U}_i^{(t-1)}, i = 1:N$; \\
 $\ten{G}^{(t)_0} \coloneqq \ten{G}^{(t-1)}$;\\
 \For{k = 1:K}
 {
 Compute $\ten{R}^{(\Delta t)}$ from \refeqn{eqn:r_delta} using $\mat{U}_i^{(t)_{k-1}}, i = 1:N$ and  $\ten{G}^{(t)_{k-1}}$ ;\\
 Compute  $\frac{\partial F^{(\Delta t)}}{\partial \mat{U}_i^{(t)_{k-1}}}$ for $i = 1:N$  from \refeqn{eqn:gradu}; \\
 Update $\mat{U}_i^{(t)_k}$ using  $\frac{\partial F^{(\Delta t)}}{\partial \mat{U}_i^{(t)_{k-1}}}$ and $\mat{U}_i^{(t)_{k-1}}$  in  \refeqn{eqn:update_u} ; \\
 Compute  $\frac{\partial F^{(\Delta t)}}{\partial \ten{G}^{(t)_{k-1}}} $   from \refeqn{eqn:gradu}; \\
 Update $\ten{G}^{(t)_k}$ using  $\ten{G}^{(t)_{k-1}}$ and   $\frac{\partial F^{(\Delta t)}}{\partial \ten{G}^{(t)_{k-1}}} $ in \refeqn{eqn:update_u}; \\
 }
 $ \mat{U}_i^{(t)} \coloneqq \mat{U}_i^{(t)_{K}}$;\\
 $ \ten{G}^{(t)} \coloneqq \ten{G}^{(t)_K}$;\\
  }
 \Output{$\mat{U}_i^{(t)}, i = 1:N, \ten{G}^{(t)}$.}
\end{algorithm}

\refalg{alg:simast} summarizes the procedure described above. The computational cost of implementing \refalg{alg:simast} depends on the update of the variables (\ref{eqn:update_u}) and the computations in (\ref{eqn:r_delta}). The cost of computing $\ten{R}^{(\Delta t)}$ is $O( \sum_i I_i M_i r_i +  |\Omega|r_1 \ldots r_N)$. The cost of performing the updates (\ref{eqn:update_u}) is $O(|\Omega| r_1 \ldots r_N + \sum_i M_i r_i)$. Overall, at every time step, the computational cost of \refalg{alg:simast} is $O(K( \sum_i I_i M_i r_i +  |\Omega|r_1 \ldots r_N))$.

\subsection*{Extension to the nonnegative case: NN-\system{}}
\label{sec:nn_siita}

We now discuss how nonnegative constraints can be incorporated into the decomposition learned by \system{}. Nonnegative constraints allow the factor of the tensor to be interpretable. 

We denote \system{} with nonnegative constraints with NN-\system{}. At every time step $t$ in the multi-aspect streaming setting, we seek to learn the following decomposition:
\begin{equation}
\label{eqn:nn_tucker}
 \underset{\mat{U}_i \in \mathbb{R}_{+}^{M_i \times r_i},  i = 1:N}{\min_{\ten{G} \in \mathbb{R}_{+}^{r_1 \times \ldots \times r_N}}}  F(\ten{X}^{(t)}, \mathcal{A}^{(t)}, \ten{G}, \{\mat{U}_i\}_{i=1:N}),
\end{equation}
where $F(\cdot)$ is as given in \eqref{eqn:F_U_G}.

We employ a projected gradient descent based algorithm for solving the optimization problem in \eqref{eqn:nn_tucker}. We follow the same incremental update scheme discussed in \refalg{alg:simast}, however we use a projection operator defined below for updating the variables. For NN-\system{}, \eqref{eqn:update_u} is replaced with
\begin{equation*}\label{eqn:update_u_nn}
\begin{array}{lll}
 \mat{U}_i^{(t)} = \Pi_{+}[\mat{U}_i^{(t-1)} - \gamma \frac{\partial F^{(\Delta t)}}{\partial \mat{U}_i^{(t-1)}}], i = 1:N \\
  \ten{G}^{(t)} =  \Pi_{+}[\ten{G}^{(t-1)} - \gamma \frac{\partial F^{(\Delta t)}}{\partial \ten{G}^{(t-1)}}],
\end{array}
\end{equation*}
where $\Pi_{+}$ is the element-wise projection operator defined as 
\begin{equation*}
 \Pi_{+}[x_i] = 
 \begin{cases}
  x_i, & \text{if}\ x_i > 0\\
  0, & \text{otherwise}.
 \end{cases}
\end{equation*}
The projection operator maps a point back to the feasible region ensuring that the factor matrices and the core tensor are always nonnegative with iterations.

\section{Experiments}
\label{sec:exp}

We evaluate \system{} against other state-of-the-art baselines in two dynamic settings viz., (1) multi-aspect streaming setting (\refsec{sec:mast}), and (2) traditional streaming setting (\refsec{sec:olstec}). We then evaluate effectiveness of \system{} in the non-streaming batch setting (\refsec{sec:static}). We analyze the effect of different types of side information in \refsec{sec:ablation}. Finally, we evaluate the performance of NN-\system{} in the unsupervised setting in \refsec{sec:nnsiita_exp}.

\textbf{Datasets}: Datasets used in the experiments are summarized in \reftbl{tbl:mast_data}. {\bf MovieLens 100K} \cite{Harper2015} is a standard movie recommendation dataset.
\textbf{YELP} is a downsampled version of the YELP(Full) dataset \cite{Jeon2016}. The YELP(Full) review dataset consists of 70K (user) $\times$ 15K (business) $\times$ 108 (year-month) tensor, and a side information matrix of size 15K (business) $\times$ 68 (city). We select a subset of this dataset for comparisons as the considered baselines algorithms cannot scale to the full dataset. We note that \system{}, our proposed method, doesn't have such scalability concerns. In \refsec{sec:ablation}, we show that \system{} scales to datasets of much larger sizes. 
In order to create YELP out of YELP(Full), we select the top frequent 1000 users and top 1000 frequent businesses and create the corresponding tensor and side information matrix. 
After the sampling, we obtain a tensor of size 1000 (user) $ \times$ 992 (business) $\times$ 93 (year-month) and a side information matrix of dimensions 992 (business) $ \times $ 56 (city). 


\begin{table}[t]
 \centering 
 \small
 \caption{\label{tbl:mast_data}Summary of datasets used in the paper. The starting size and increment size given in the table are for Multi-Aspect Streaming setting. For Streaming setting, the tensor grows in the third dimension, one slice at every time step.}
 \begin{tabular}{ccc}
  \toprule
   & MovieLens 100K & YELP \\
   \midrule
   Modes & user $\times$ movie $\times$ week & user $\times$ business $\times$ year-month \\
    \midrule
   Tensor Size & 943$\times$1682$\times$31& 1000$\times$992$\times$93 \\
   \midrule 
   Starting size & 19$\times$34$\times$2&  20$\times$20$\times$2\\
 \midrule  
   Increment step &19, 34, 1 & 20, 20, 2\\
 \midrule  
   Sideinfo matrix & 1682 (movie) $\times$ 19 (genre) & 992 (business) $\times$ 56 (city) \\
   \bottomrule
 \end{tabular}
\end{table}


\subsection{Multi-Aspect Streaming Setting}
\label{sec:mast}

\begin{table}[t]
 \centering
 \small
 \caption{\label{tbl:rmse_mast}Test RMSE (lower is better) averaged across all the time steps in the multi-aspect streaming tensor sequence setting (Definition 4) for MAST and \system{}. \system{}, the proposed method,  outperforms MAST for all the datasets. \refsec{sec:mast} provides more details.}
 \begin{tabular}{ccccc}
 \toprule
 Dataset & Missing\% & Rank & MAST & \system{} \\
  \midrule
    
  \multirow{9}{35pt}{MovieLens 100K}  &\multirow{3}{5pt}{20\%} &	3	&  1.60& {\bf 1.23}\\
				&	& 5		& 1.53	& 1.29\\
				&	& 10		& 1.48	&2.49\\
				\cline{2-5}
				&\multirow{3}{5pt}{50\%} &	3	& 1.74	& {\bf 1.28}\\
				&	& 5	& 1.75	& 1.29\\
				&	& 10	& 1.64	& 2.55\\				
				\cline{2-5}
				&\multirow{3}{5pt}{80\%} &	3	& 2.03	& {\bf 1.59}\\
				&	& 5	& 1.98	&1.61\\
				&	& 10	& 2.02	&2.96\\							
\midrule
				\multirow{9}{30pt}{YELP} & \multirow{3}{5pt}{20\%}	& 	3 	&	1.90& {\bf 1.43}  \\
				  &	&	5	& 1.92	& 1.54\\
				&	&	10	& 1.93	&4.03\\
				\cline{2-5}
				& \multirow{3}{5pt}{50\%}	& 	3 	& 1.94	& {\bf 1.51} \\
				  &	&	5	& 	1.94& 1.67\\
				&	&	10	& 1.96	& 4.04\\
				\cline{2-5}
				& \multirow{3}{5pt}{80\%}	& 	3 	&1.97	& 1.71 \\
				  &	&	5	& 	1.97 & {\bf 1.61}\\
				&	&	10	& 1.97	& 3.49\\
 \bottomrule
 \end{tabular}
\end{table}

\begin{figure}
\centering
\begin{tabular}{cc}
\noindent \begin{minipage}[b]{0.5\hsize}
\centering
\includegraphics[scale=0.25]{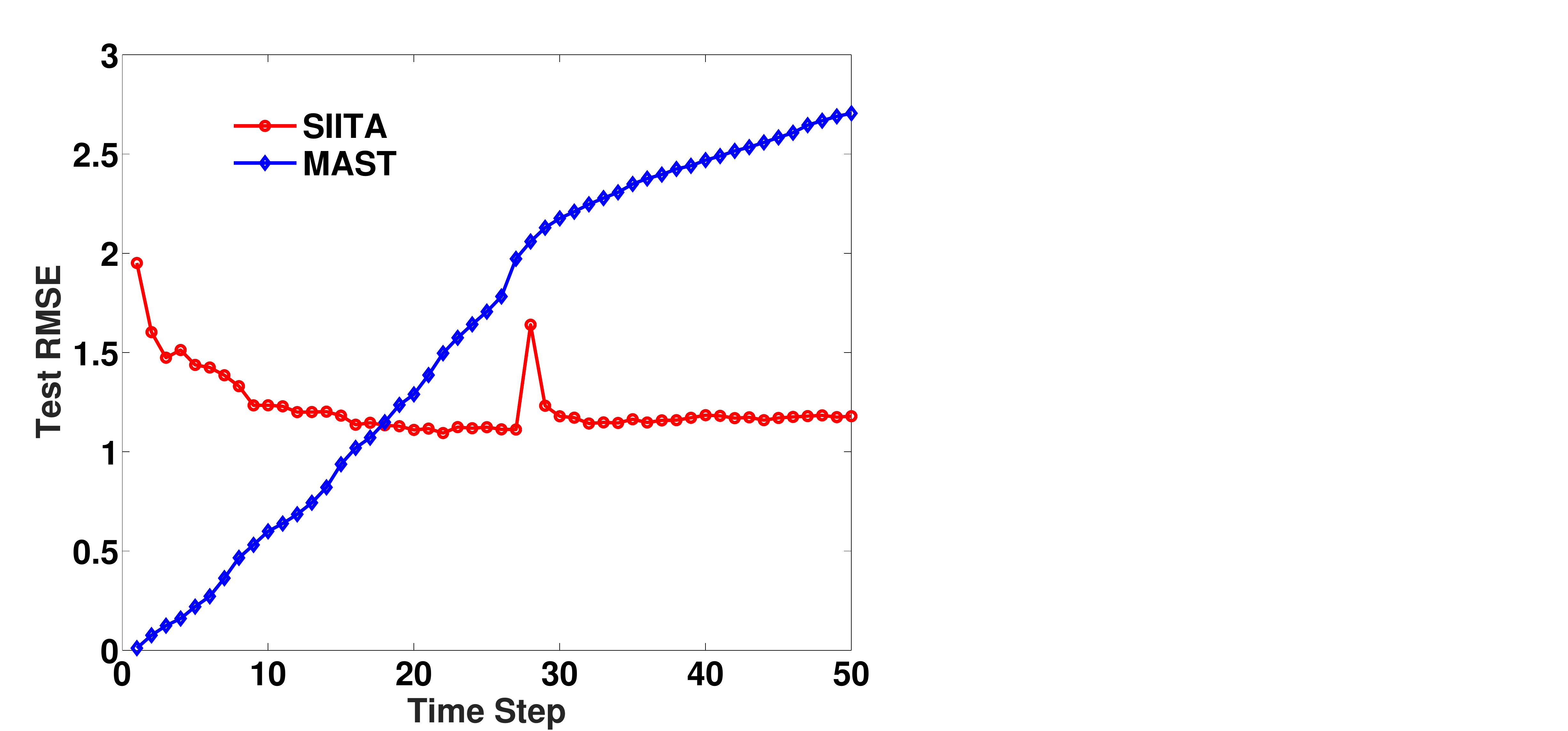}\\
{\small (a) MovieLens 100K \\ (20\% Missing)}
\end{minipage}
\begin{minipage}[b]{0.5\hsize}
\centering
\includegraphics[scale=0.25]{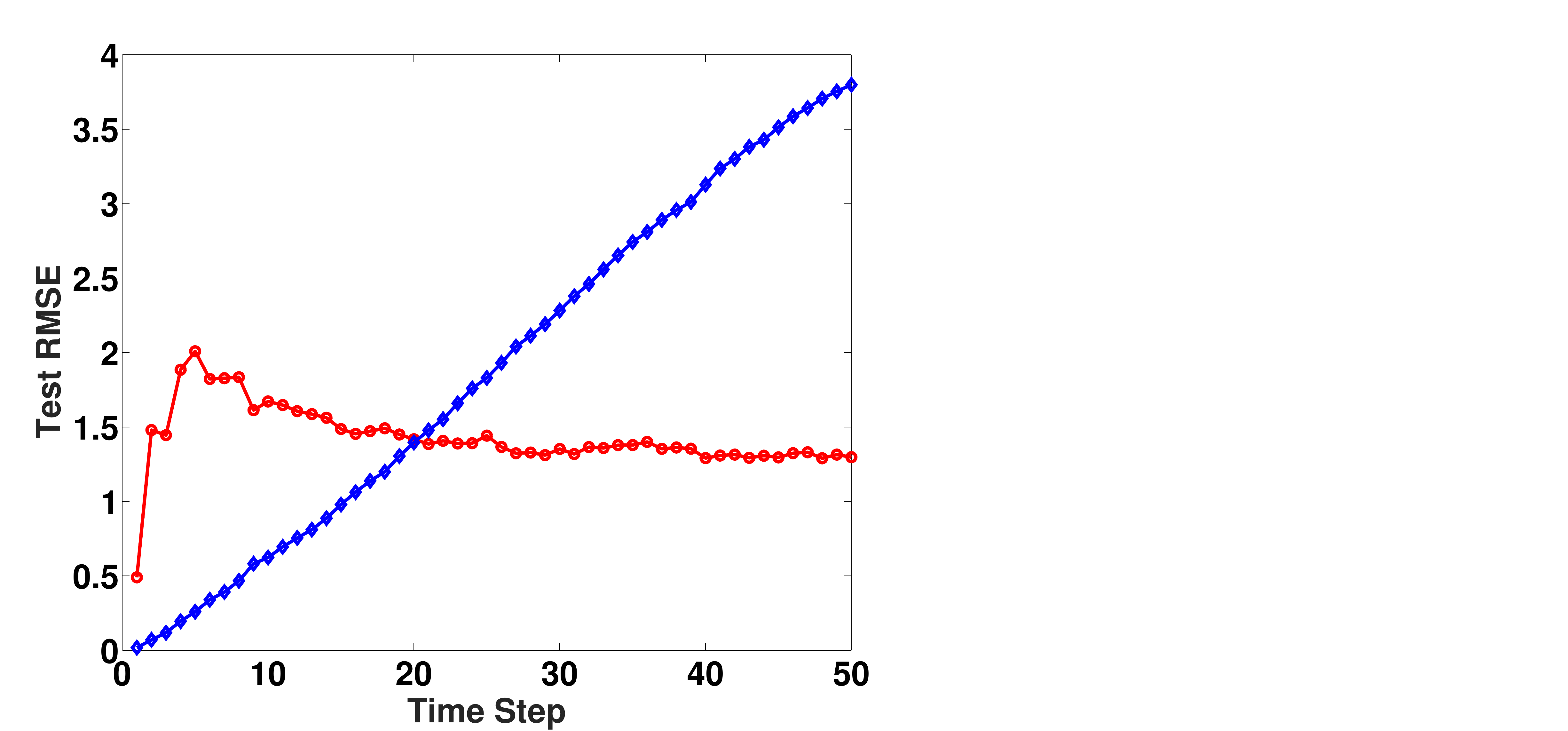}\\
{\small (b) YELP \\ (20\% Missing)}
\end{minipage}
\end{tabular}
\caption{\label{fig:mast_rmse}Evolution of test RMSE of MAST and \system{} with each time step. For both the datasets, \system{} attains a stable performance after a few time steps, while the performance of MAST degrades with every time step. Refer to \refsec{sec:mast} for more details.}
\end{figure}

\begin{figure}
\centering
\begin{tabular}{cc}
\noindent \begin{minipage}[b]{0.5\hsize}
\centering
\includegraphics[scale=0.25]{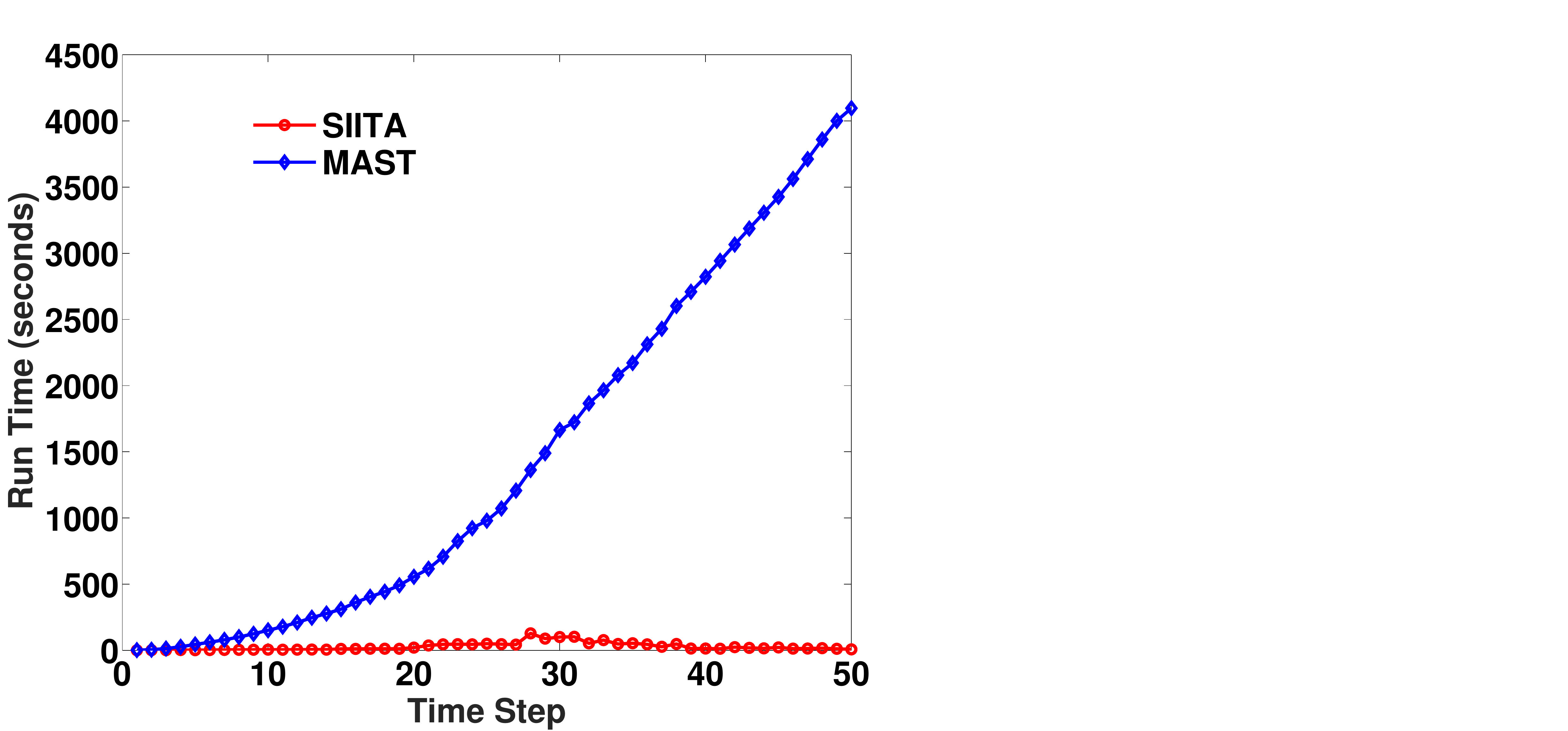}\\
{\small (a) MovieLens 100K \\ (20\% Missing)}
\end{minipage}
\begin{minipage}[b]{0.5\hsize}
\centering
\includegraphics[scale=0.25]{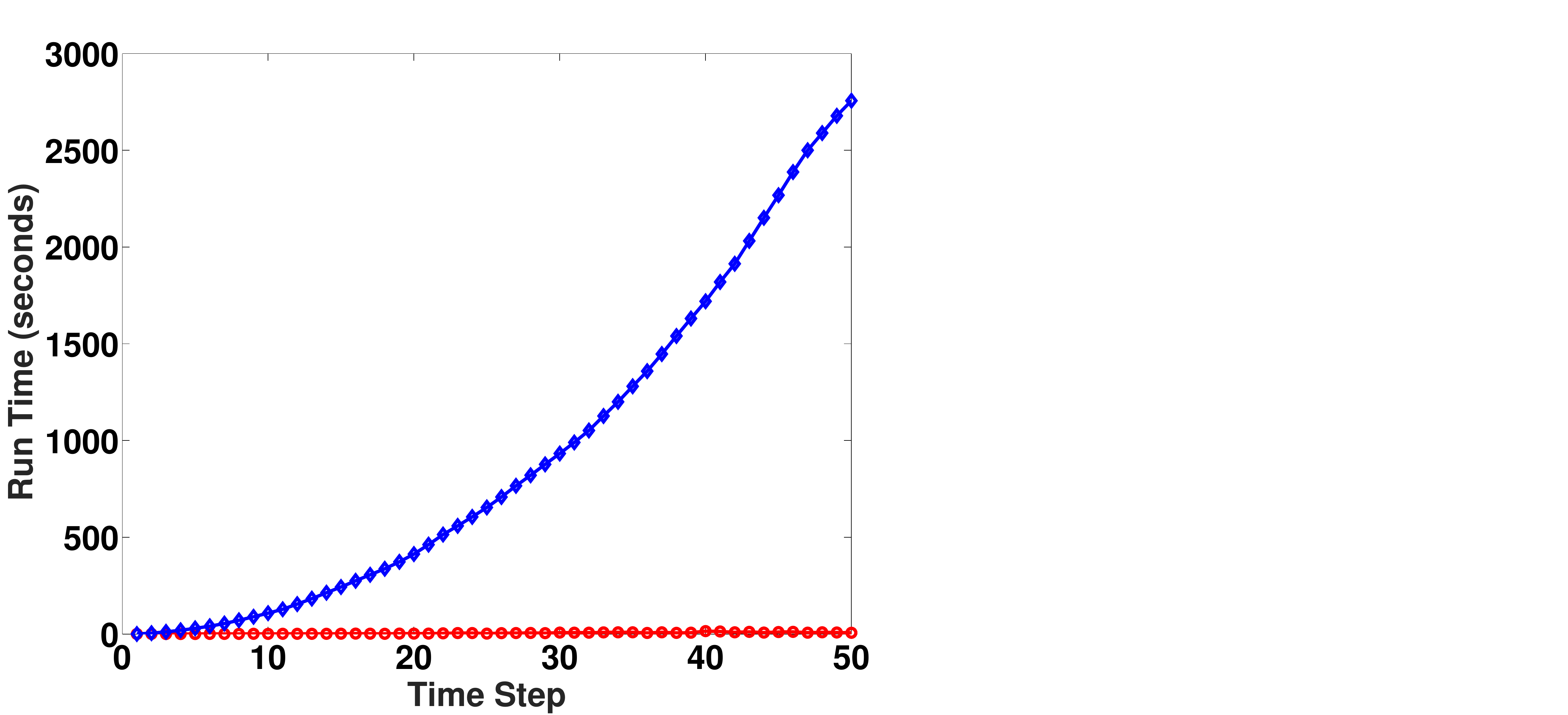}\\
{\small (b) YELP  \\ (20\% Missing)}
\end{minipage}
\end{tabular}
\caption{\label{fig:mast_runtime}Runtime comparison between MAST and \system{} at every time step. 
\system{} is significantly faster than MAST. Refer to \refsec{sec:mast} for more details.}
\end{figure}

We first analyze the model in the multi-aspect streaming setting, for which we consider MAST \cite{Song2017} as a state-of-the-art baseline. \\
{\bf MAST \cite{Song2017}}: MAST is a dynamic low-rank tensor completion algorithm, which enforces nuclear norm regularization on the decomposition matrices of CP. A tensor-based Alternating Direction Method of Multipliers is used for solving the optimization problem.

We experiment with the MovieLens 100K and YELP datasets. Since the third mode is time in both the datasets, i.e., (week) in MovieLens 100K and (year-month) in YELP, one way to simulate the multi-aspect streaming sequence (Definition 3) is by considering 
every slice in third-mode as one time step in the sequence, and letting the tensor grow along other two modes with every time step, similar to the ladder structure given in \cite[Section ~3.3]{Song2017}. Note that this is different from the traditional streaming setting, where the tensor only grows in time mode while the other two modes remain fixed.
In contrast, in the multi-aspect setting here, there can be new users joining the system within the same month but on different days or different movies getting released on different days in the same week etc. Therefore in our simulations, we consider the third mode as any normal mode and generate a more general multi-aspect streaming tensor sequence, the details are given in \reftbl{tbl:mast_data}. The parameters for MAST are set based on the guidelines provided in \cite[~Section 4.3]{Song2017}.

We compute the root mean square error on test data (test RMSE; lower is better) at every time step and report the test RMSE averaged across all the time steps in \reftbl{tbl:rmse_mast}. 
We perform experiments on multiple train-test splits for each dataset. We vary the test percentage, denoted by {\it Missing\%} in \reftbl{tbl:rmse_mast}, and the rank of decomposition, denoted by {\it Rank} for both the datasets. For every (Missing\%, Rank) combination, we run both models on ten random train-test splits and report the average. 
For \system{}, Rank = $r$ in \reftbl{tbl:rmse_mast} represents the Tucker-rank $(r, r, r)$.

In \reftbl{tbl:rmse_mast}, the proposed \system{} achieves better results than MAST. \reffig{fig:mast_rmse} shows the plots for test RMSE at every time step. Since  \system{} handles the sparsity in the data effectively, as a result \system{} is significantly faster than MAST, which can be seen from \reffig{fig:mast_runtime}. Overall, we find that \system{}, the proposed method, is more effective and faster compared to MAST in the multi-aspect streaming setting. 

%

 \subsection{Streaming Setting}
 \label{sec:olstec}

 \begin{figure}
\centering
\begin{tabular}{cc}
\begin{minipage}[b]{0.5\hsize}
\centering
\includegraphics[scale=0.25]{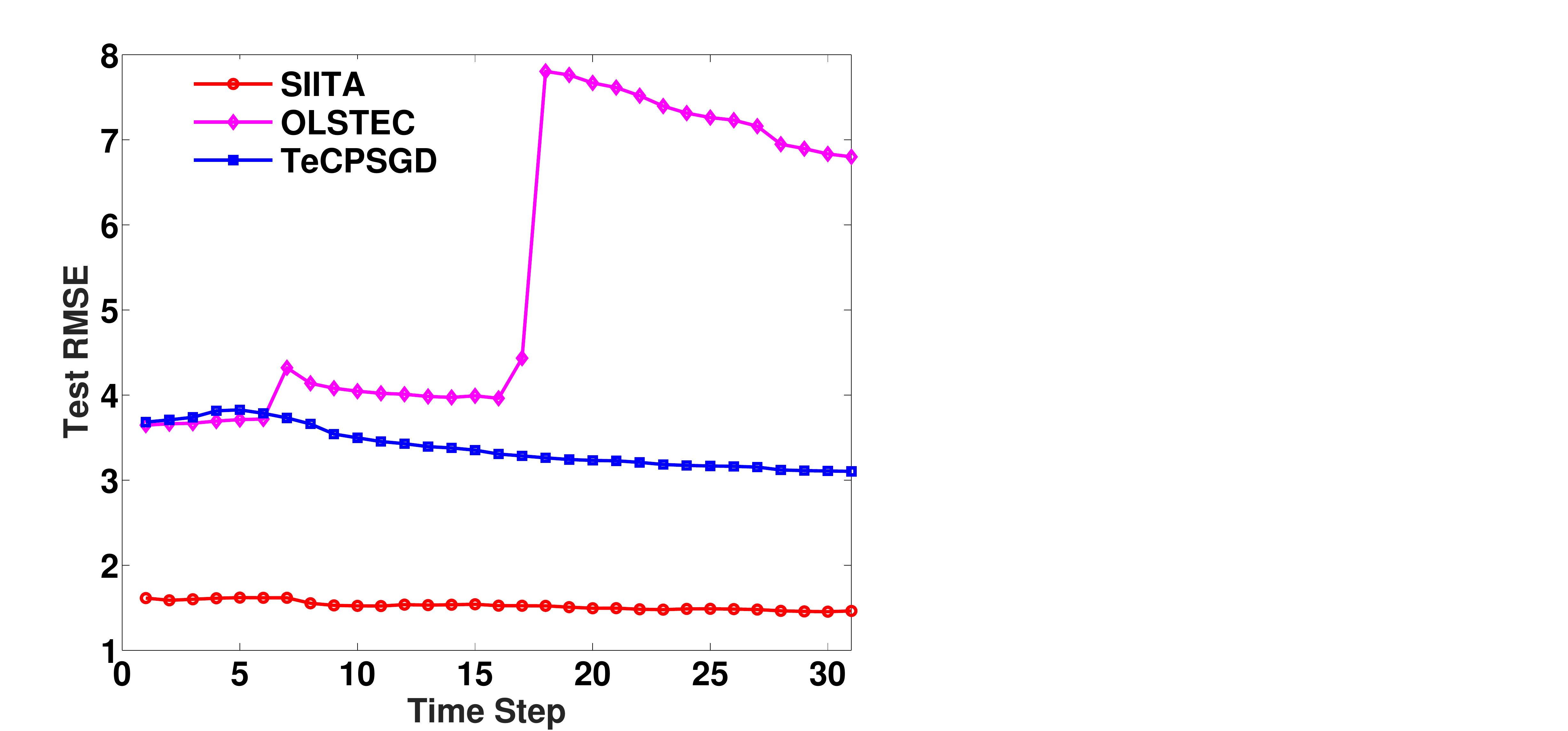}\\
{\small (b) MovieLens 100K \\ (20\% Missing)}
\end{minipage}
\noindent \begin{minipage}[b]{0.5\hsize}
\centering
\includegraphics[scale=0.254]{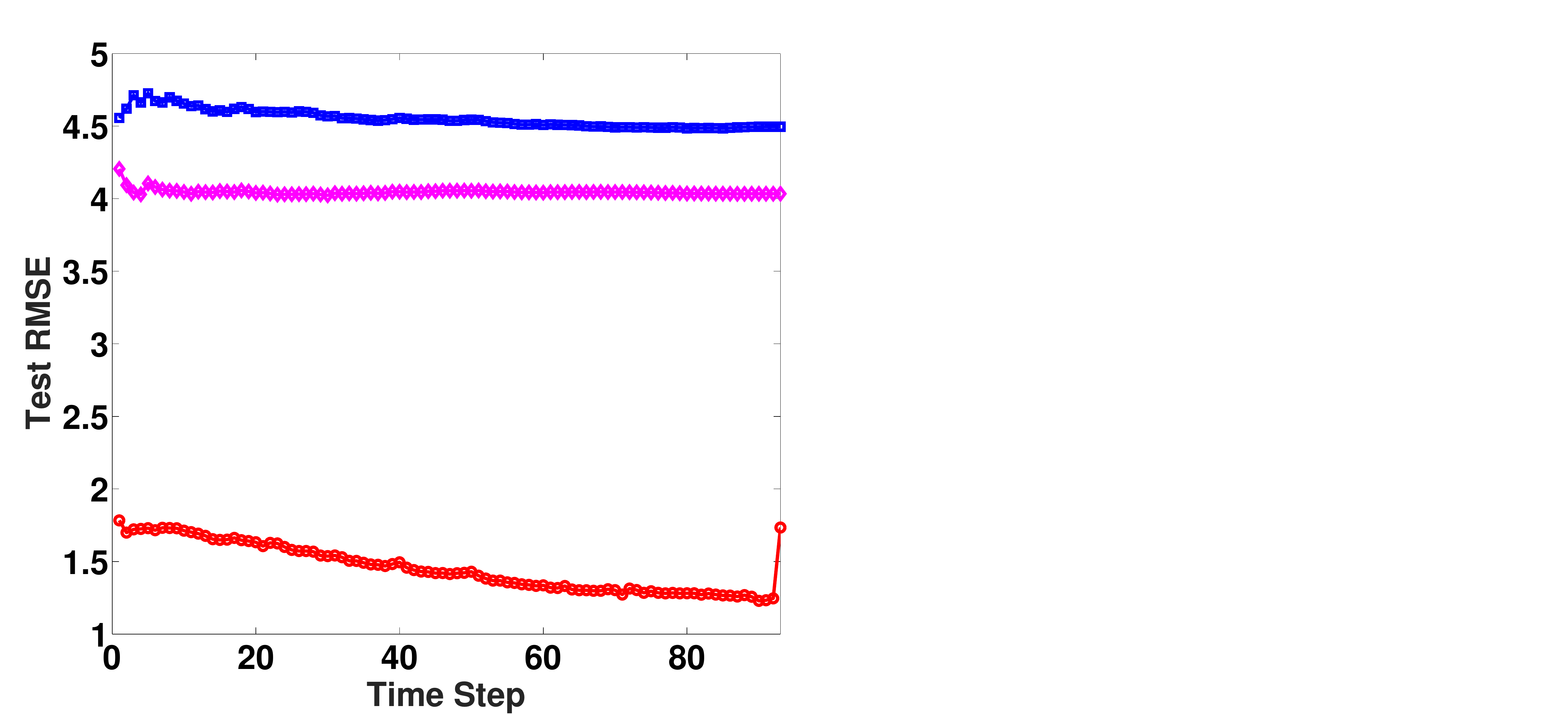}\\
{\small (a) YELP \\ (20\% Missing)}
\end{minipage}
\end{tabular}
\caption{\label{fig:olstec_rmse} Evolution of Test RMSE of TeCPSGD,  OLSTEC and \system{} with each time step. In both datasets, \system{} performs significantly better than the baseline algorithms in the pure streaming setting. 
Refer to \refsec{sec:olstec} for more details.}
\end{figure}

\begin{figure}
\centering
\begin{tabular}{cc}
\begin{minipage}[b]{0.5\hsize}
\centering
\includegraphics[scale=0.25]{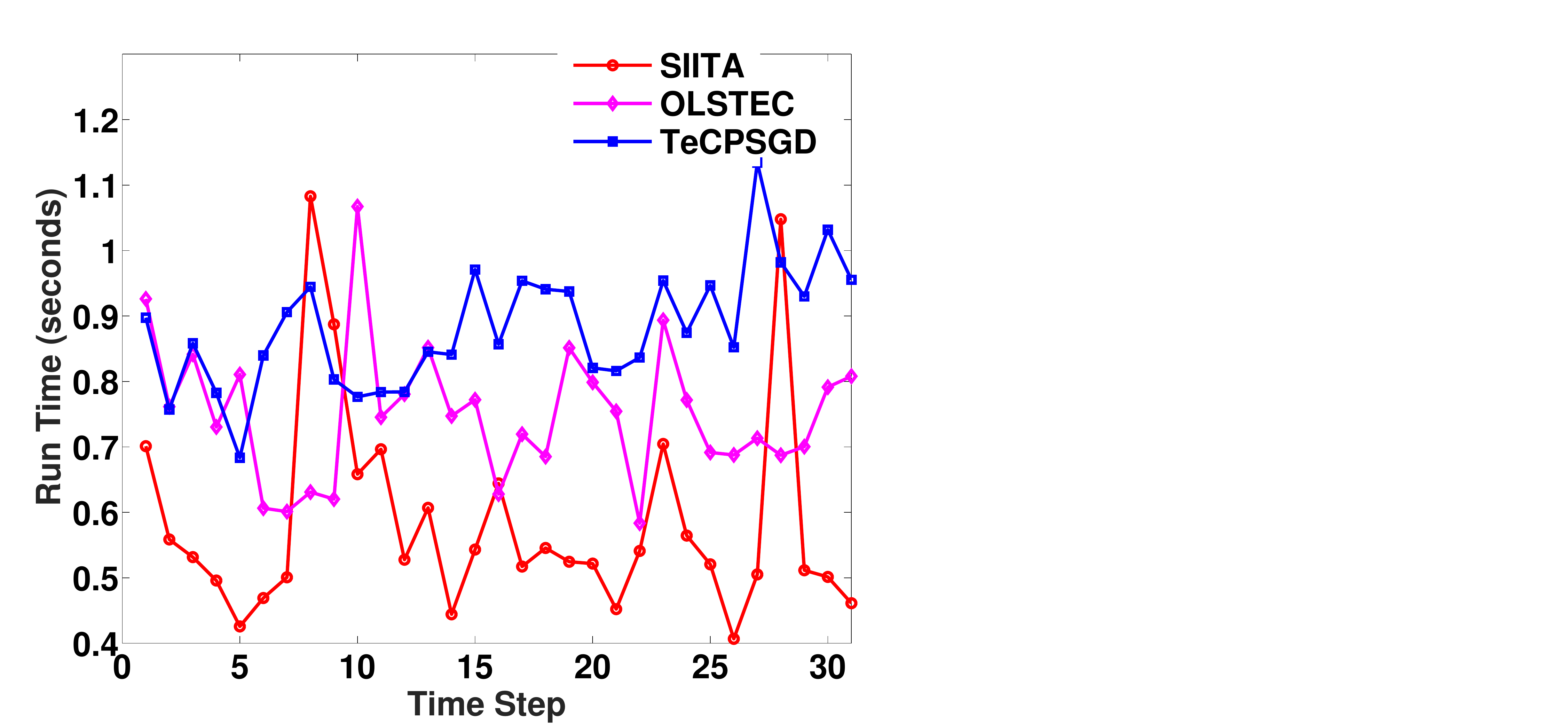}\\
{\small (b)  MovieLens 100K \\ (20\% Missing)}
\end{minipage}
\noindent \begin{minipage}[b]{0.5\hsize}
\centering
\includegraphics[scale=0.254]{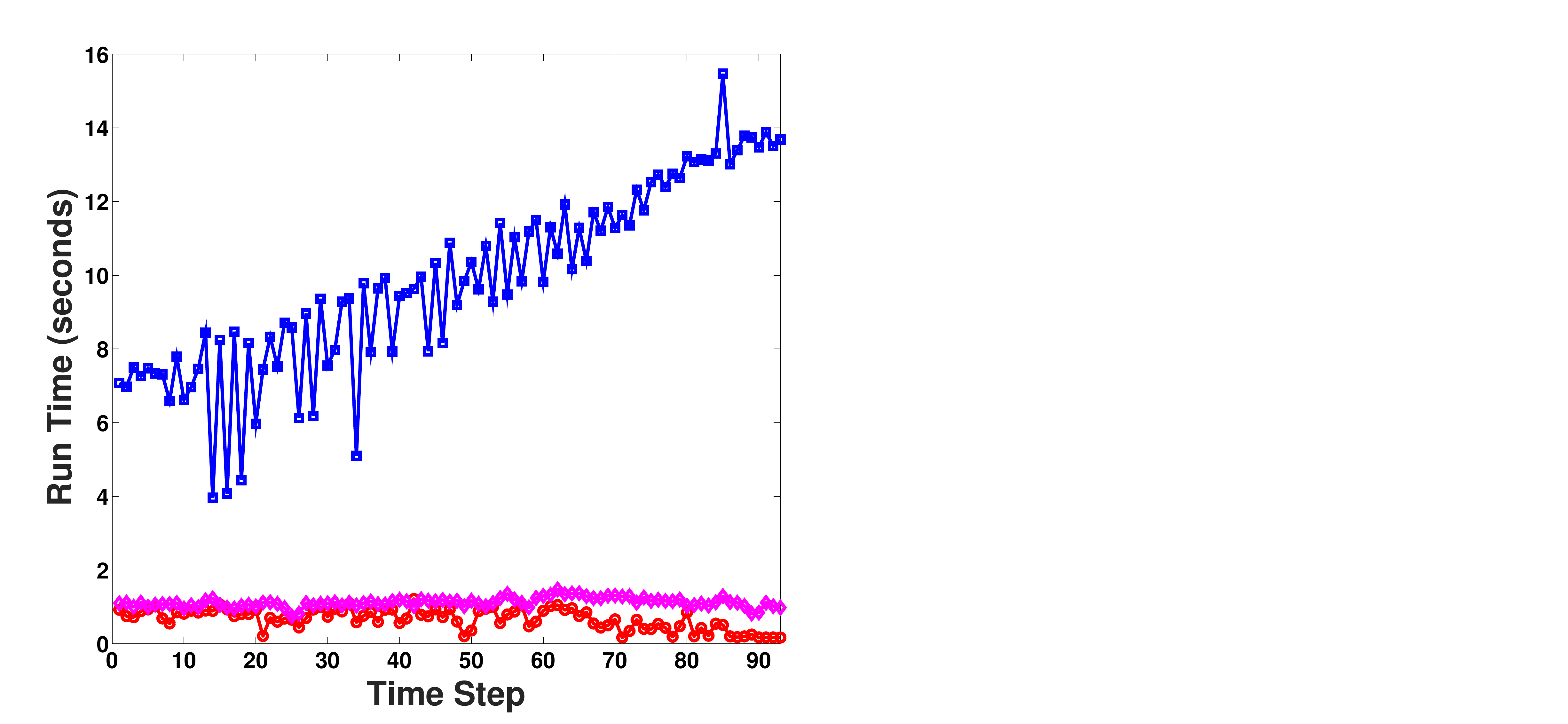}\\
{\small (a)  YELP  \\ (20\% Missing)}
\end{minipage}
\end{tabular}
\caption{\label{fig:olstec_runtime}Runtime comparison between TeCPSGD, OLSTEC and \system{}. \system{} is able to exploit sparsity in the data and is much faster. Refer to \refsec{sec:olstec} for more details.}
\end{figure}

\begin{table}[t]
 \centering
 \small
 \caption{Test RMSE averaged across all the time steps in the streaming setting for  TeCPSGD, OLSTEC, a state-of-the-art streaming tensor completion algorithm, and SIITA.  \system{} outperforms the baseline algorithms significantly. See \refsec{sec:olstec} for more details. \label{tbl:rmse_streaming}}
 \begin{tabular}{cccccc}
 \toprule
 Dataset & Missing\% & Rank & TeCPSGD & OLSTEC & \system{} \\
  \midrule
  
  \multirow{9}{35pt}{MovieLens 100K}  &\multirow{3}{5pt}{20\%} &	3	& 3.39	& 5.46 &{\bf 1.53}\\
				&	& 5		& 3.35	& 4.65& 1.54\\
				&	& 10		& 3.19	& 4.96& 1.71\\
				\cline{2-6}
				&\multirow{3}{5pt}{50\%} &	3	& 3.55	& 8.39& {\bf 1.63}\\
				&	& 5	& 3.40	& 6.73& 1.64\\
				&	& 10	& 3.23	&3.66 &1.73\\				
				\cline{2-6}
				&\multirow{3}{5pt}{80\%} &	3	& 3.78	&3.82 & 1.79\\
				&	& 5	& 3.77	&  3.80& {\bf 1.75}\\
				&	& 10	& 3.84	& 4.34& 2.47\\							
\midrule
				\multirow{9}{30pt}{YELP} & \multirow{3}{5pt}{20\%}	& 	3 	& 4.55	& 4.04 & {\bf 1.45}\\
				  &	&	5	& 4.79	& 4.04& 1.59\\
				&	&	10	& 5.17	& 4.03& 2.85\\
				\cline{2-6}
				& \multirow{3}{5pt}{50\%}	& 	3 	& 4.67	&  4.03& {\bf 1.55}\\
				  &	&	5	& 5.03	& 4.03& 1.67\\
				&	&	10	& 5.25	& 4.03 & 2.69\\
				\cline{2-6}
				& \multirow{3}{5pt}{80\%}	& 	3 	& 4.99	&  4.02& {\bf 1.73}\\
				  &	&	5	& 5.17	& 4.02& 1.78\\
				&	&	10	& 5.31	& 4.01& 2.62\\
 \bottomrule
 \end{tabular}
\end{table}
 

In this section, we simulate the pure streaming setting by letting the tensor grow only in the third mode at every time step. The number of time steps for each dataset in this setting is the dimension of the third mode, i.e., 31 for MovieLens 100K and 93 for YELP.
We compare the performance of \system{} with TeCPSGD and OLSTEC algorithms in the streaming setting.\\
{\bf TeCPSGD \cite{Mardani2015}:} TeCPSGD is an online Stochastic Gradient Descent based algorithm for recovering missing data in streaming tensors. This algorithm is based on PARAFAC decomposition. TeCPSGD is the first proper tensor completion algorithm in the dynamic setting.\\
{\bf OLSTEC \cite{Kasai2016}:} OLSTEC   is an online tensor tracking algorithm for partially observed data streams corrupted by noise. OLSTEC is a second order stochastic gradient descent algorithm based on CP decomposition exploiting recursive least squares. OLSTEC is the state-of-the-art for streaming tensor completion. 

We report test RMSE, averaged across all time steps, for both MovieLens 100K and YELP datasets. Similar to the multi-aspect streaming setting, we run all the algorithms for multiple train-test splits. For each split, we run all the algorithms with different ranks. For every (Missing\%, Rank) combination, we run all the algorithms on ten random train-test splits and report the average. \system{} significantly outperforms all the baselines in this setting, as shown in \reftbl{tbl:rmse_streaming}. Figure \ref{fig:olstec_rmse} shows the average test RMSE of every algorithm at every time step. From \reffig{fig:olstec_runtime} it can be seen that \system{} takes much less time compared to other algorithms. The spikes in the plots suggest that the particular slices are relatively less sparse. 

\subsection{Batch Setting}
\label{sec:static}

\begin{table}[t]
 \centering
 \small
 \caption{Mean Test RMSE across multiple train-test splits in the Batch setting. \system{} achieves lower test RMSE on both the datasets compared to AirCP, a state-of-the-art algorithm for this setting. Refer to \refsec{sec:static} for details. \label{tbl:rmse_static}}
 \begin{tabular}{ccccc}
  \toprule
  Dataset & Missing\% & Rank & AirCP & \system{} \\
  \midrule
  
  \multirow{9}{35pt}{MovieLens 100K}  &\multirow{3}{5pt}{20\%} &	3	& 3.351	& {\bf1.534}\\
				&	& 5	&	3.687 & 1.678\\
				&	& 10	&	3.797 &2.791\\
				\cline{2-5}
				&\multirow{3}{5pt}{50\%} &	3	& 3.303	& {\bf 1.580}\\
				&	& 5	&	3.711& 1.585\\
				&	& 10	&	3.894& 2.449\\				
				\cline{2-5}
				&\multirow{3}{5pt}{80\%} &	3	& 3.883	& {\bf 1.554}\\
				&	& 5	&	3.997 & 1.654\\
				&	& 10	&	3.791 & 3.979\\							
\midrule
				\multirow{9}{30pt}{YELP} & \multirow{3}{5pt}{20\%}	& 	3 	& 	1.094 & {\bf1.052} \\
				  &	&	5	&	1.086 & 1.056\\
				&	&	10	&	1.077 & 1.181\\
				\cline{2-5}
				& \multirow{3}{5pt}{50\%}	& 	3 	& 	1.096 & 1.097 \\
				  &	&	5	&	1.095	& {\bf1.059}\\
				&	&	10	&	1.719 & 1.599\\
				\cline{2-5}
				& \multirow{3}{5pt}{80\%}	& 	3 	& 1.219	 & 1.199 \\
				  &	&	5	&	{\bf 1.118} & 1.156\\
				&	&	10	&	2.210 & 2.153\\

  \bottomrule
 \end{tabular}
\end{table}

Even though our primary focus is on proposing an algorithm for the multi-aspect streaming setting, \system{} can be run as a tensor completion algorithm with side information in the batch (i.e., non streaming) setting. To run in batch setting, we set $K=1$ in \refalg{alg:simast} and run for multiple passes over the data. In this setting, AirCP \cite{Ge2016} is the current state-of-the-art algorithm which is also capable of handling side information. We consider AirCP as the baseline in this section. The main focus of this setting is to demonstrate that \system{} incorporates the side information effectively.
~\\
{\bf AirCP \cite{Ge2016}}: AirCP is a CP based tensor completion algorithm proposed for recovering the spatio-temporal dynamics of online memes. This algorithm incorporates auxiliary information from memes, locations and times. An alternative direction method of multipliers (ADMM) based algorithm is employed for solving the optimization. AirCP expects the side information matrices to be similarity matrices and takes input the Laplacian of the similarity matrices. However, in the datasets we experiment with, the side information is available as feature matrices. Therefore, we consider the covariance matrices $\mat{A}_i\mat{A}_i^\top$ as similarity matrices.

We run both algorithms till convergence and report test RMSE. For each dataset, we experiment with different levels of test set sizes, 
and for each such level, we run our experiments on 10 random splits. We report the mean test RMSE per train-test percentage split. We run our experiments with multiple ranks of factorization. Results are shown in \reftbl{tbl:rmse_static}, where we observe that \system{} achieves better results. Note that the rank for \system{} is the Tucker rank, i.e., rank = 3. This implies a factorization rank of (3, 3, 3) for \system{}. 

\textbf{Remark:} Since all the baselines considered for various settings are CP based, we only compare for CP tensor rank. From Tables \ref{tbl:rmse_mast}, \ref{tbl:rmse_streaming} and \ref{tbl:rmse_static} it can be seen that the performance suffers for rank = 10. However, when we run \system{} with a rank = (10, 10, 2) we achieve a lower test RMSE.

\subsection{Analyzing Merits of Side Information}
\label{sec:ablation}

\begin{table}[t]
 \centering
 \small
 \caption{\label{tbl:mast_no_si} Test RMSE averaged across multiple train-test splits in the Multi-Aspect Streaming setting, analyzing the merits of side information. See \refsec{sec:ablation} for more details.}
 \begin{tabular}{ccccc}
 \toprule
 Dataset & Missing\% & Rank & \system{} (w/o SI) & \system{} \\
  \midrule
    
  \multirow{9}{35pt}{MovieLens 100K}  &\multirow{3}{5pt}{20\%} &	3	&  {\bf 1.19}&  1.23\\
				&	& 5		& {\bf 1.19}	& 1.29\\
				&	& 10		& 2.69	&2.49\\
				\cline{2-5}
				&\multirow{3}{5pt}{50\%} &	3	& {\bf 1.25}	& 1.28\\
				&	& 5	& {\bf 1.25}	& 1.29\\
				&	& 10	& 3.28	& 2.55\\				
				\cline{2-5}
				&\multirow{3}{5pt}{80\%} &	3	& 1.45	&  1.59\\
				&	& 5	& {\bf 1.42}	&1.61\\
				&	& 10	& 2.11	&2.96\\							
\midrule
				\multirow{9}{30pt}{YELP} & \multirow{3}{5pt}{20\%}	& 	3 	&	1.44& {\bf 1.43}  \\
				  &	&	5	& 1.48	& 1.54\\
				&	&	10	& 3.90	&4.03\\
				\cline{2-5}
				& \multirow{3}{5pt}{50\%}	& 	3 	& 1.57	& {\bf 1.51} \\
				  &	&	5	& 	1.62& 1.67\\
				&	&	10	& 5.48	& 4.04\\
				\cline{2-5}
				& \multirow{3}{5pt}{80\%}	& 	3 	&1.75	& 1.71 \\
				  &	&	5	& 	1.67 & {\bf 1.61}\\
				&	&	10	& 5.28	& 3.49\\
 \bottomrule
 \end{tabular}
\end{table}

\begin{table}[t]
 \centering
 \small
 \caption{\label{tbl:streaming_no_si} Test RMSE averaged across multiple train-test splits in the streaming setting, analyzing the merits of side information. See \refsec{sec:ablation} for more details.}
 \begin{tabular}{ccccc}
 \toprule
 Dataset & Missing\% & Rank & \system{} (w/o SI) & \system{} \\
  \midrule
    
  \multirow{9}{35pt}{MovieLens 100K}  &\multirow{3}{5pt}{20\%} &	3	& {\bf 1.46} &  1.53\\
				&	& 5		& 1.53	& 1.54\\
				&	& 10		& 1.55	&1.71\\
				\cline{2-5}
				&\multirow{3}{5pt}{50\%} &	3	& 1.58	& 1.63\\
				&	& 5	& 1.67	& 1.64\\
				&	& 10	& {\bf 1.56}	& 1.73\\				
				\cline{2-5}
				&\multirow{3}{5pt}{80\%} &	3	& 1.76	& 1.79\\
				&	& 5	& {\bf 1.74}	&1.75\\
				&	& 10	& 2.31	&2.47\\							
\midrule
				\multirow{9}{30pt}{YELP} & \multirow{3}{5pt}{20\%}	& 	3 	& 1.46	& {\bf 1.45}  \\
				  &	&	5	& 1.62	& 1.59\\
				&	&	10	& 2.82	&2.85\\
				\cline{2-5}
				& \multirow{3}{5pt}{50\%}	& 	3 	& 1.57	& {\bf 1.55} \\
				  &	&	5	& 1.69	& 1.67\\
				&	&	10	& 2.54	& 2.67\\
				\cline{2-5}
				& \multirow{3}{5pt}{80\%}	& 	3 	&1.76	& {\bf 1.73} \\
				  &	&	5	& 1.80	 & 1.78\\
				&	&	10	& 2.25	& 2.62\\
 \bottomrule
 \end{tabular}
\end{table}

\begin{figure}
\centering
\begin{tabular}{cc}
\begin{minipage}[b]{0.5\hsize}
\centering
\includegraphics[scale=0.25]{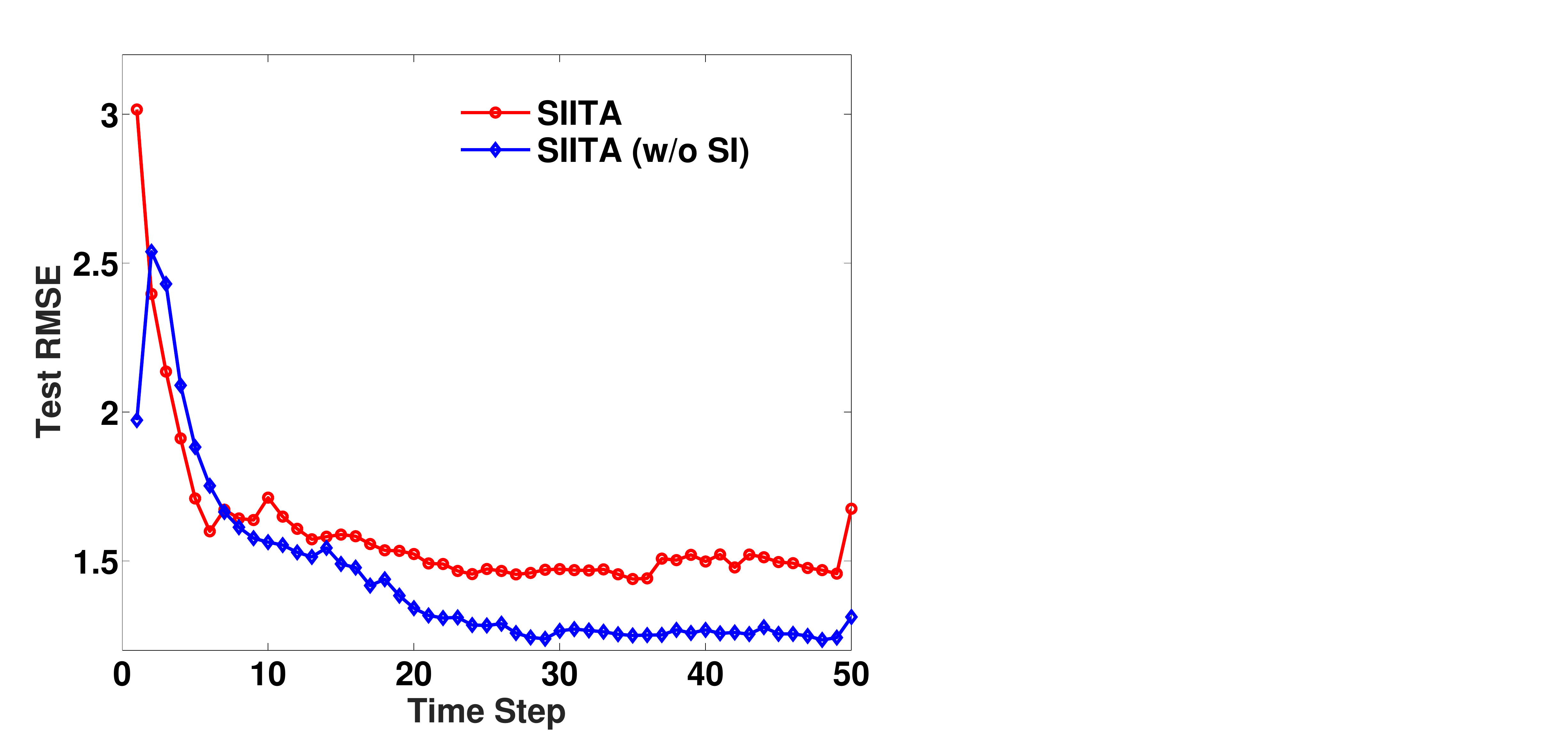}\\
{\small (b)  MovieLens 100K \\ (80\% Missing)}
\end{minipage}
\noindent \begin{minipage}[b]{0.5\hsize}
\centering
\includegraphics[scale=0.25]{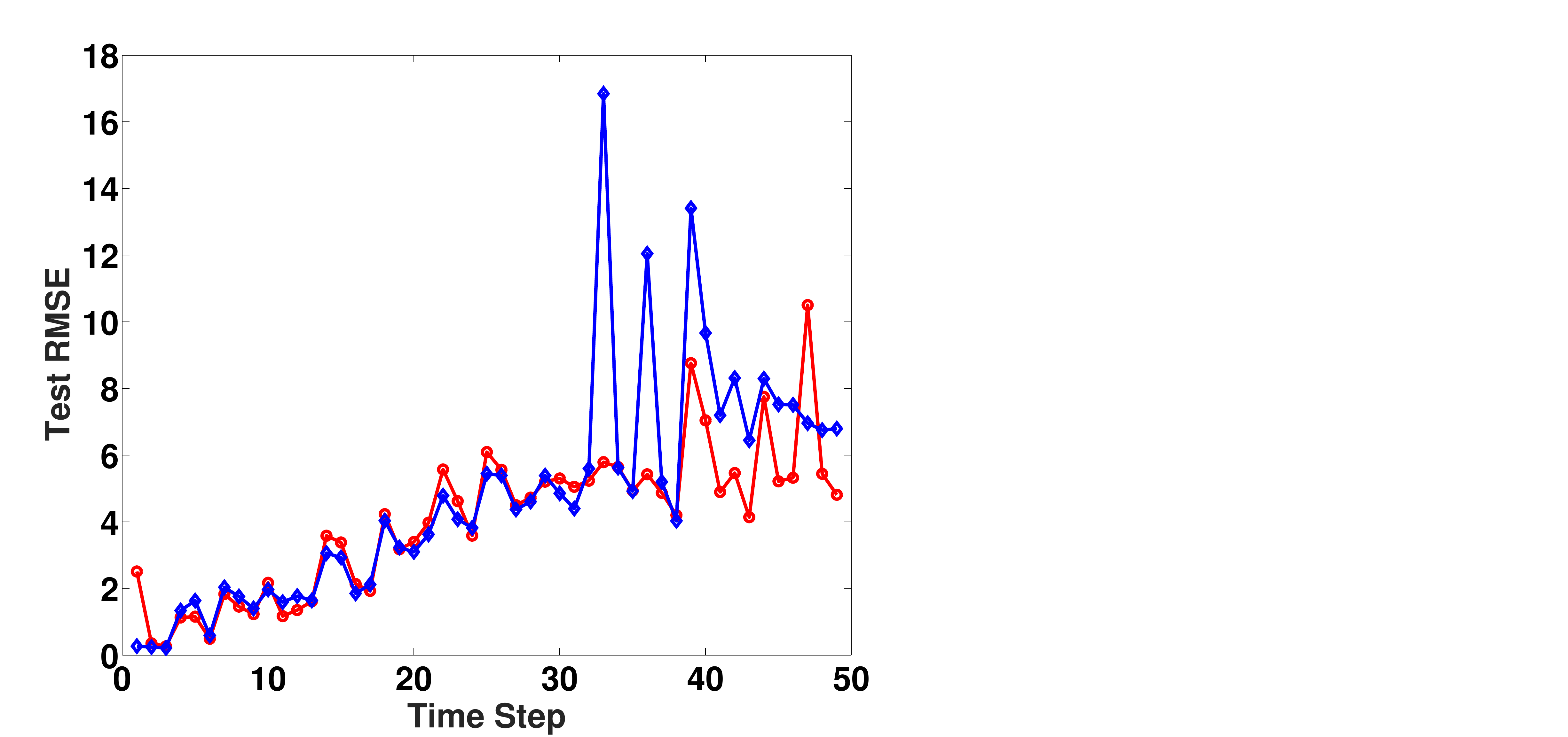}\\
{\small (a)  YELP  \\ (80\% Missing)}
\end{minipage}
\end{tabular}
\caption{\label{fig:rmse_no_si_mast}Evolution of test RMSE with every time step in the multi-aspect streaming setting for \system{} and \system{} (w/o SI). See \refsec{sec:ablation} for more details.}
\end{figure}

\begin{figure}
\centering
\begin{tabular}{cc}
\begin{minipage}[b]{0.5\hsize}
\centering
\includegraphics[scale=0.25]{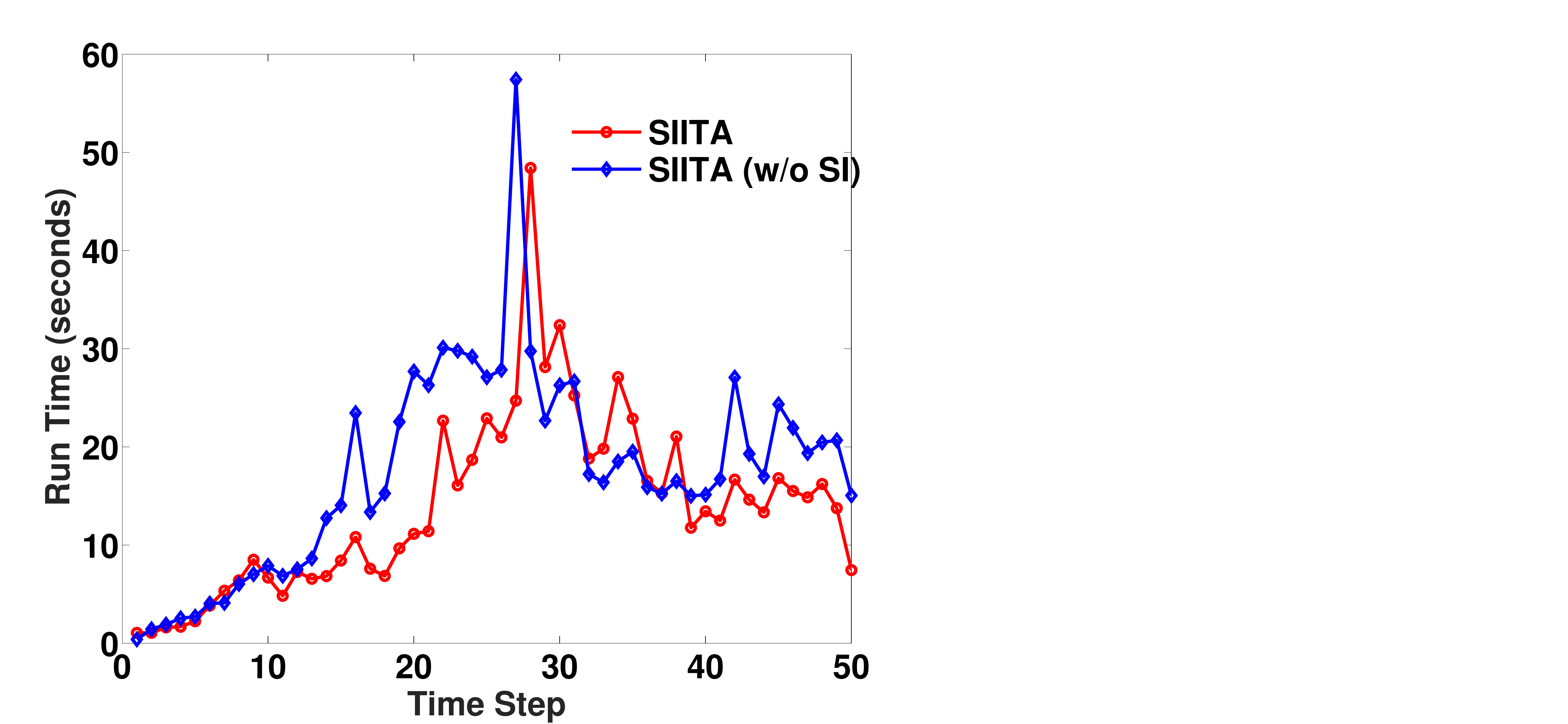}\\
{\small (b)  MovieLens 100K \\ (80\% Missing)}
\end{minipage}
\noindent \begin{minipage}[b]{0.5\hsize}
\centering
\includegraphics[scale=0.25]{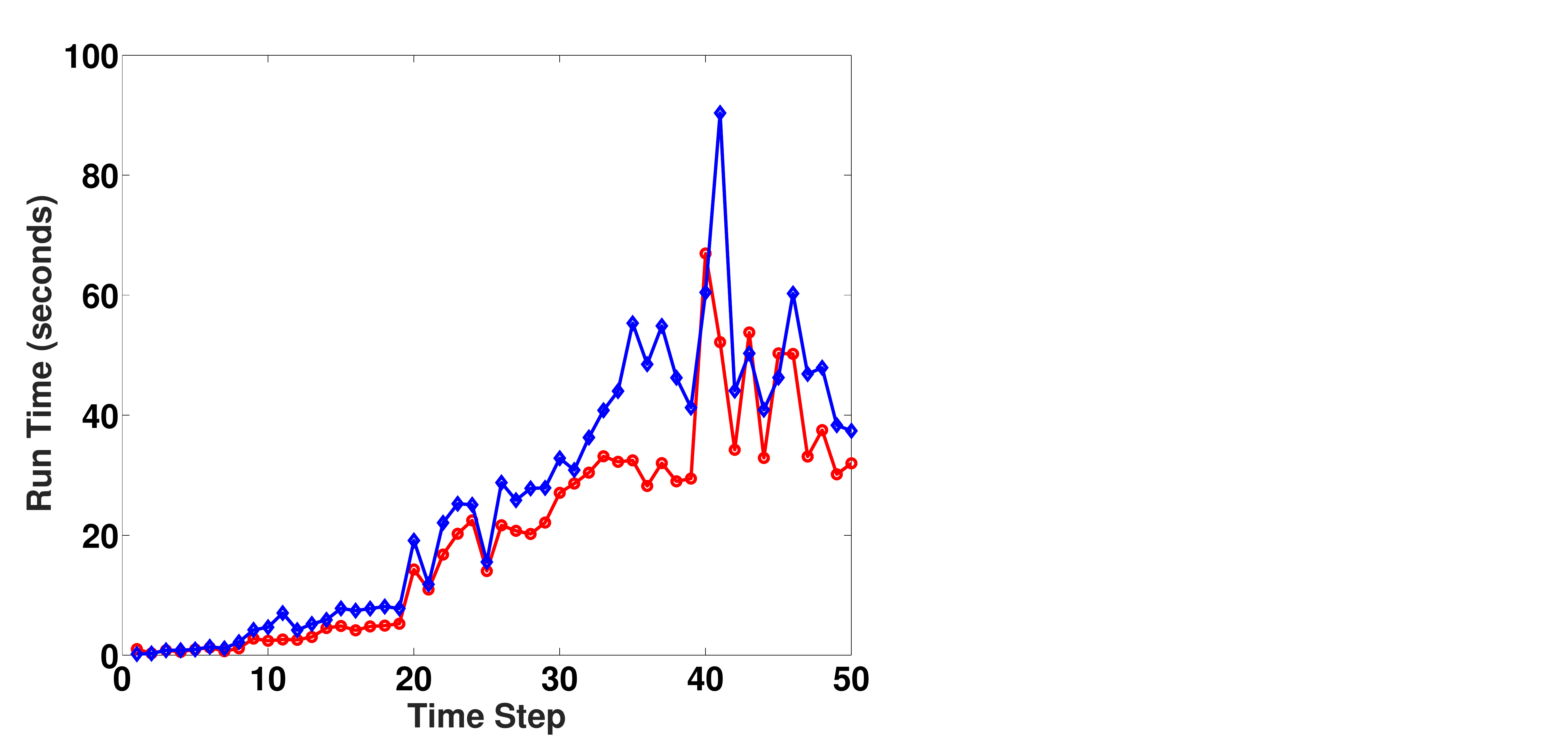}\\
{\small (a)  YELP  \\ (80\% Missing)}
\end{minipage}
\end{tabular}
\caption{\label{fig:runtime_no_si_mast}Run Time comparison between \system{} and \system{} (w/o SI) in the multi-aspect streaming setting. See \refsec{sec:ablation} for more details.}
\end{figure}

\begin{figure}
\centering
\begin{tabular}{cc}
\begin{minipage}[b]{0.5\hsize}
\centering
\includegraphics[scale=0.25]{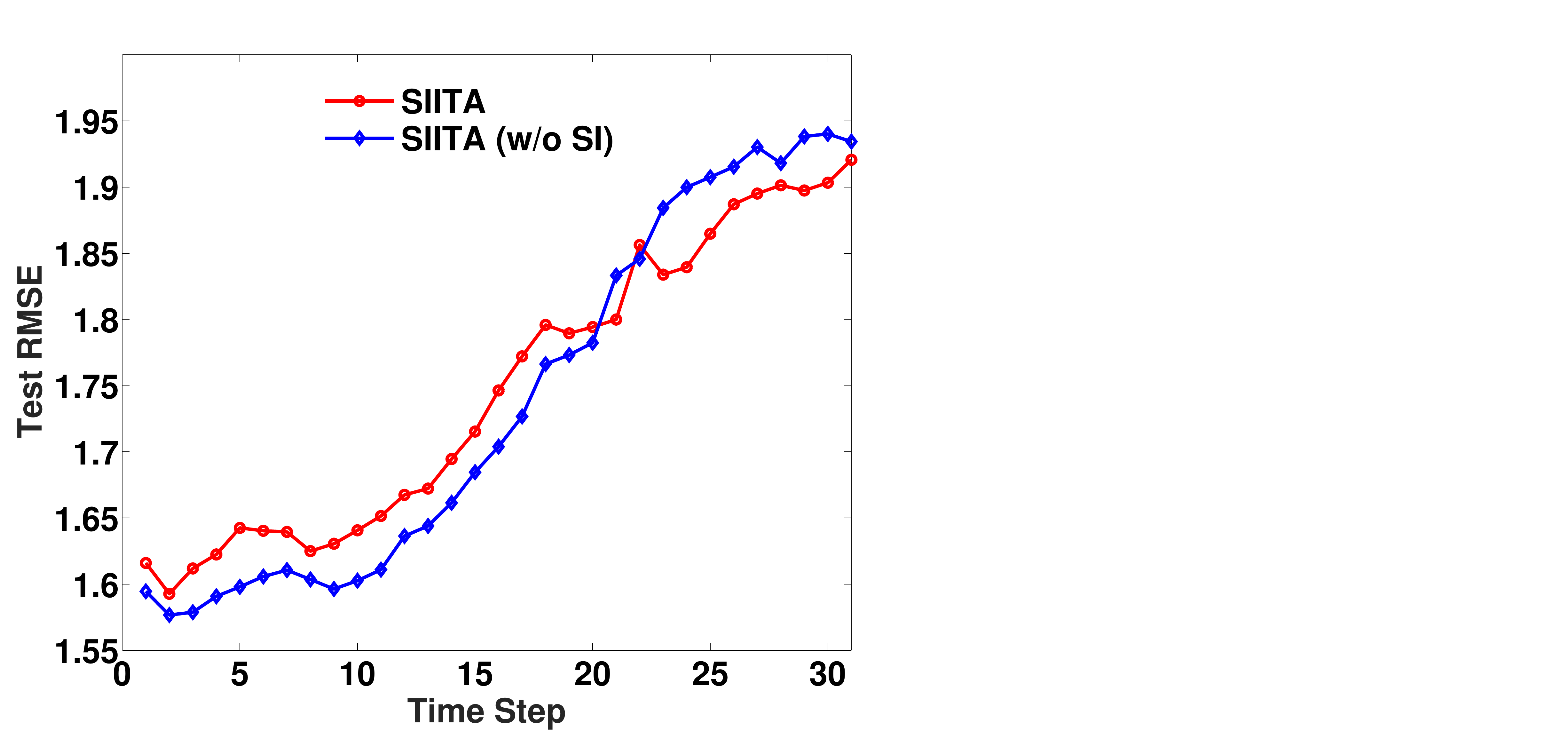}\\
{\small (b)  MovieLens 100K \\ (80\% Missing)}
\end{minipage}
\noindent \begin{minipage}[b]{0.5\hsize}
\centering
\includegraphics[scale=0.25]{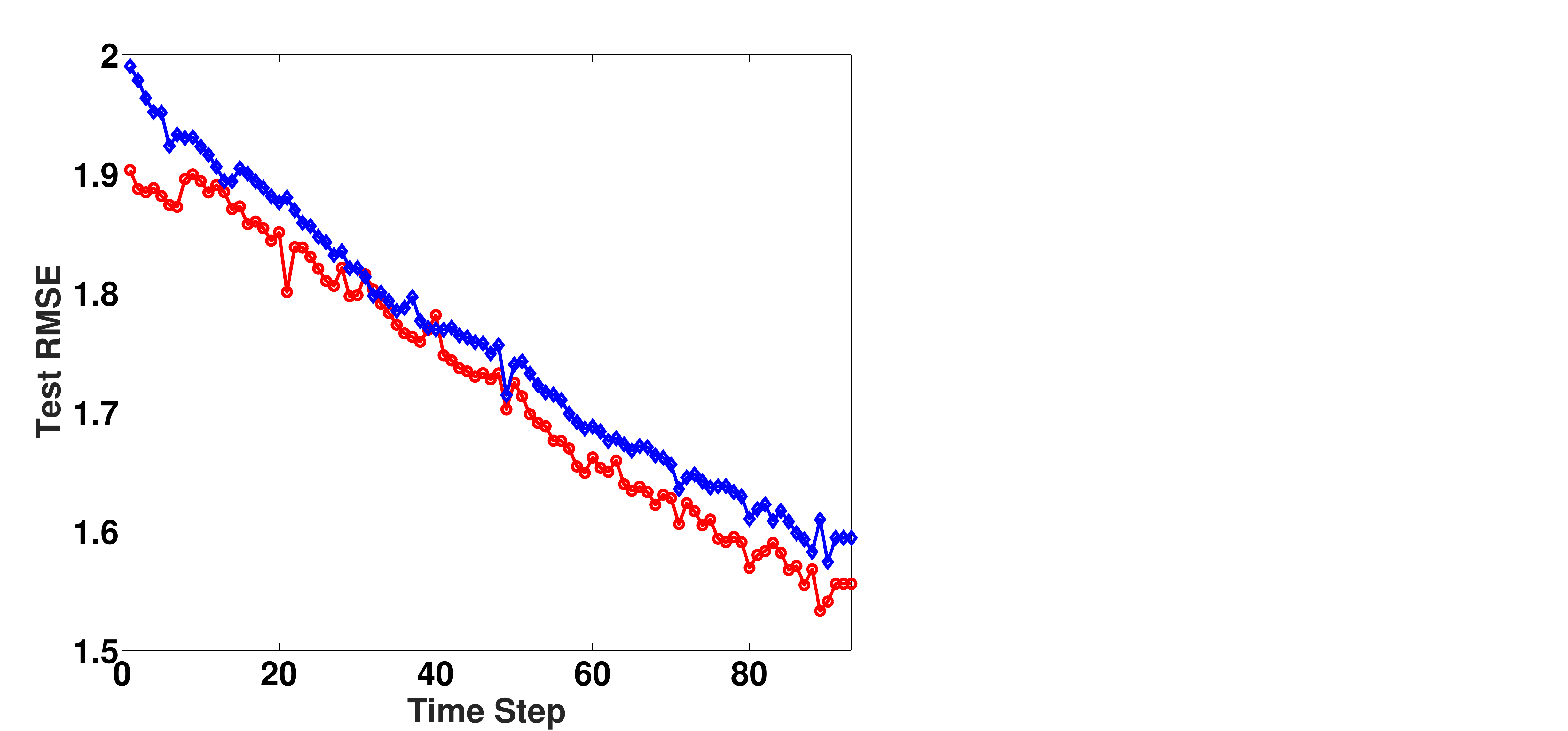}\\
{\small (a)  YELP  \\ (80\% Missing)}
\end{minipage}
\end{tabular}
\caption{\label{fig:rmse_no_si_streaming}Evolution of test RMSE with every time step in the streaming setting for \system{} and \system{}(w/o SI). See \refsec{sec:ablation} for more details.}
\end{figure}

\begin{figure}
\centering
\begin{tabular}{cc}
\begin{minipage}[b]{0.5\hsize}
\centering
\includegraphics[scale=0.25]{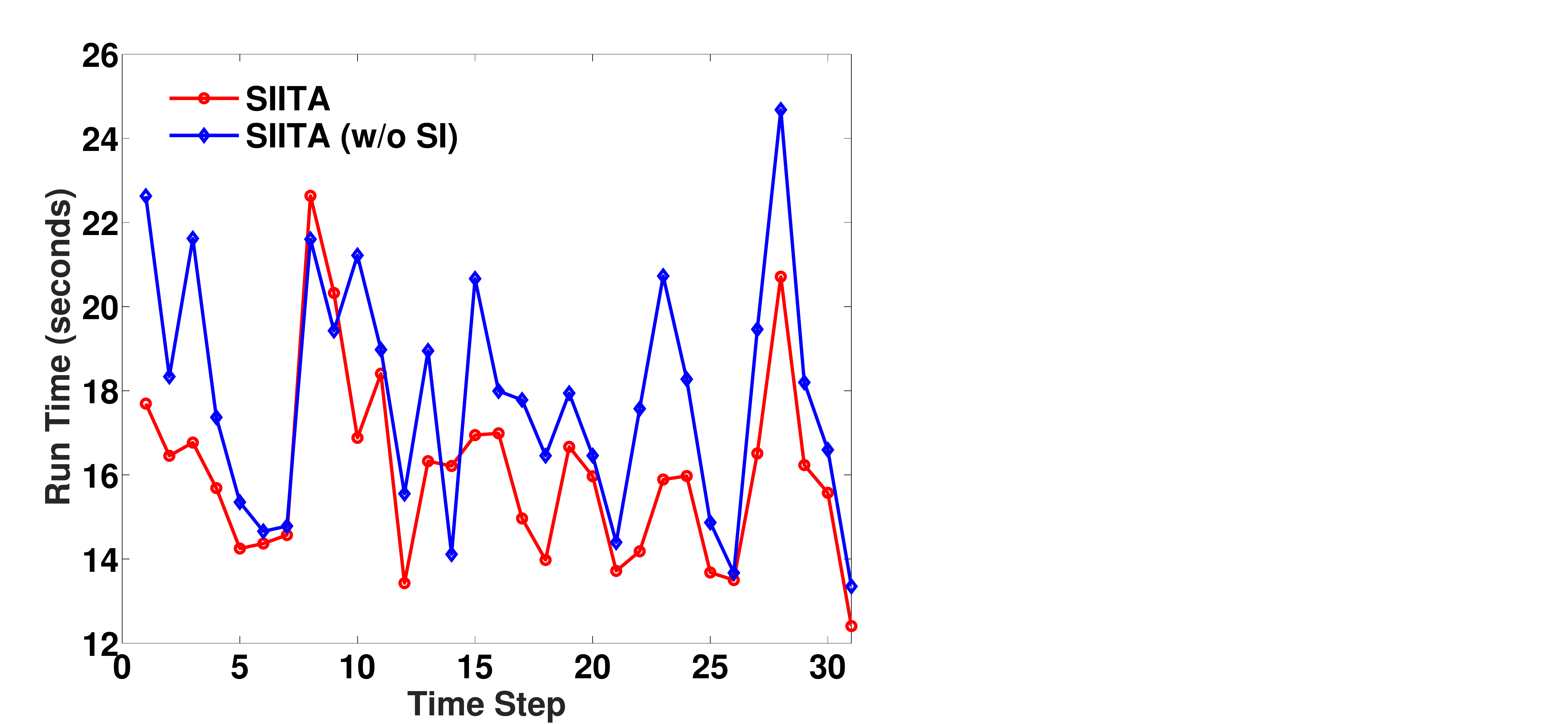}\\
{\small (b)  MovieLens 100K \\ (80\% Missing)}
\end{minipage}
\noindent \begin{minipage}[b]{0.5\hsize}
\centering
\includegraphics[scale=0.25]{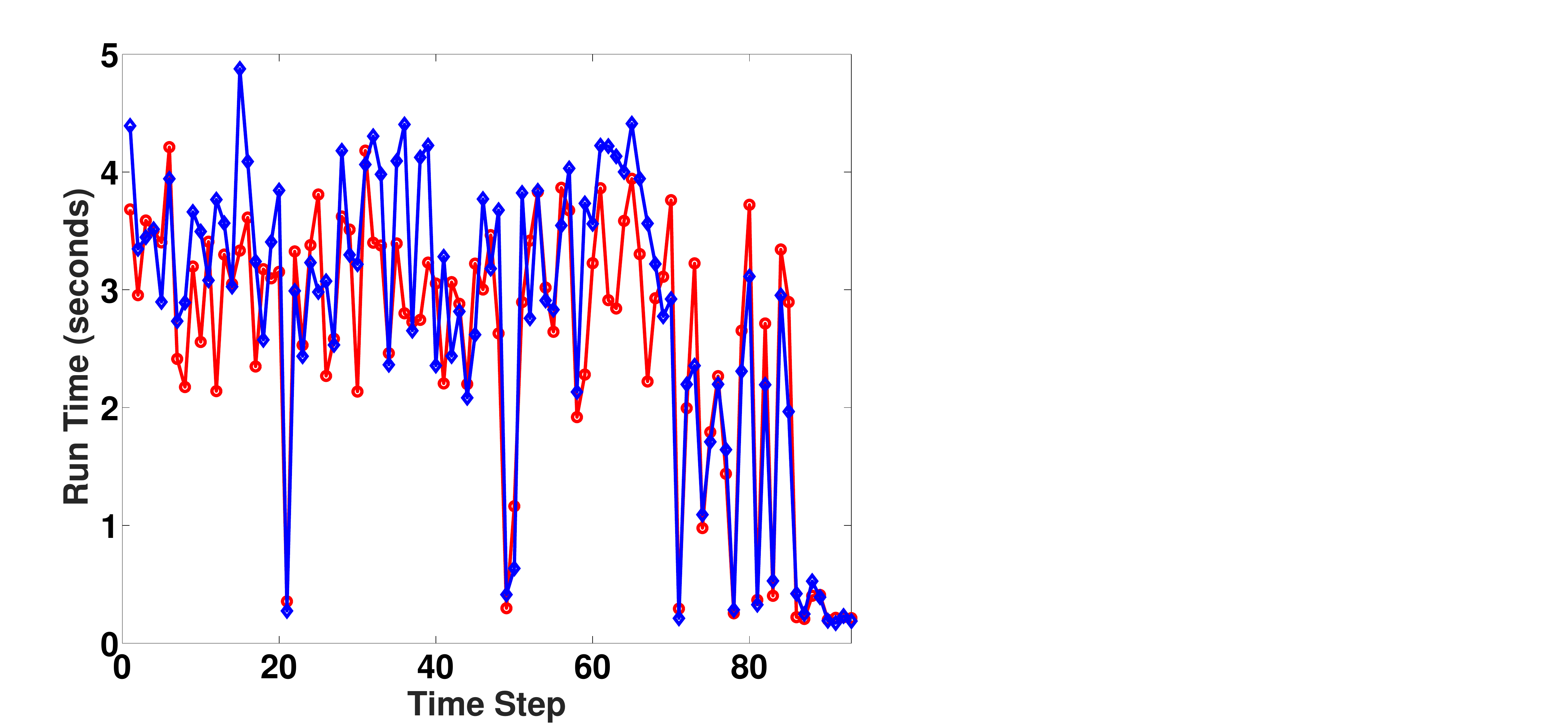}\\
{\small (a)  YELP  \\ (80\% Missing)}
\end{minipage}
\end{tabular}
\caption{\label{fig:runtime_no_si_streaming}Run Time comparison between \system{} and \system{} (w/o SI) in the Streaming setting. See \refsec{sec:ablation} for more details.}
\end{figure}

\begin{figure}
\centering
\begin{tabular}{cc}
\begin{minipage}[b]{0.5\hsize}
\centering
\includegraphics[scale=0.24]{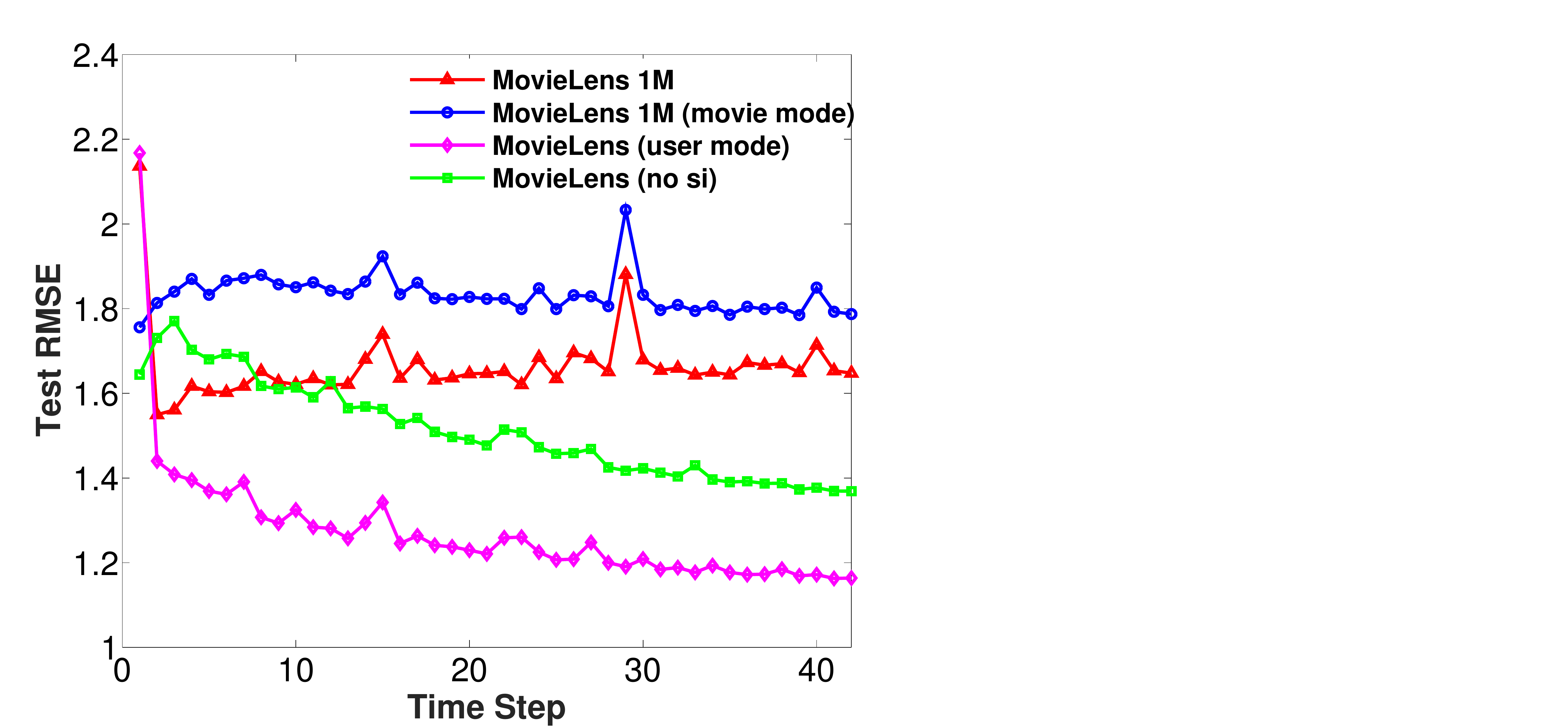}\\
{\small (a)  Test RMSE at every time step}
\end{minipage}
\noindent \begin{minipage}[b]{0.5\hsize}
\centering
\includegraphics[scale=0.25]{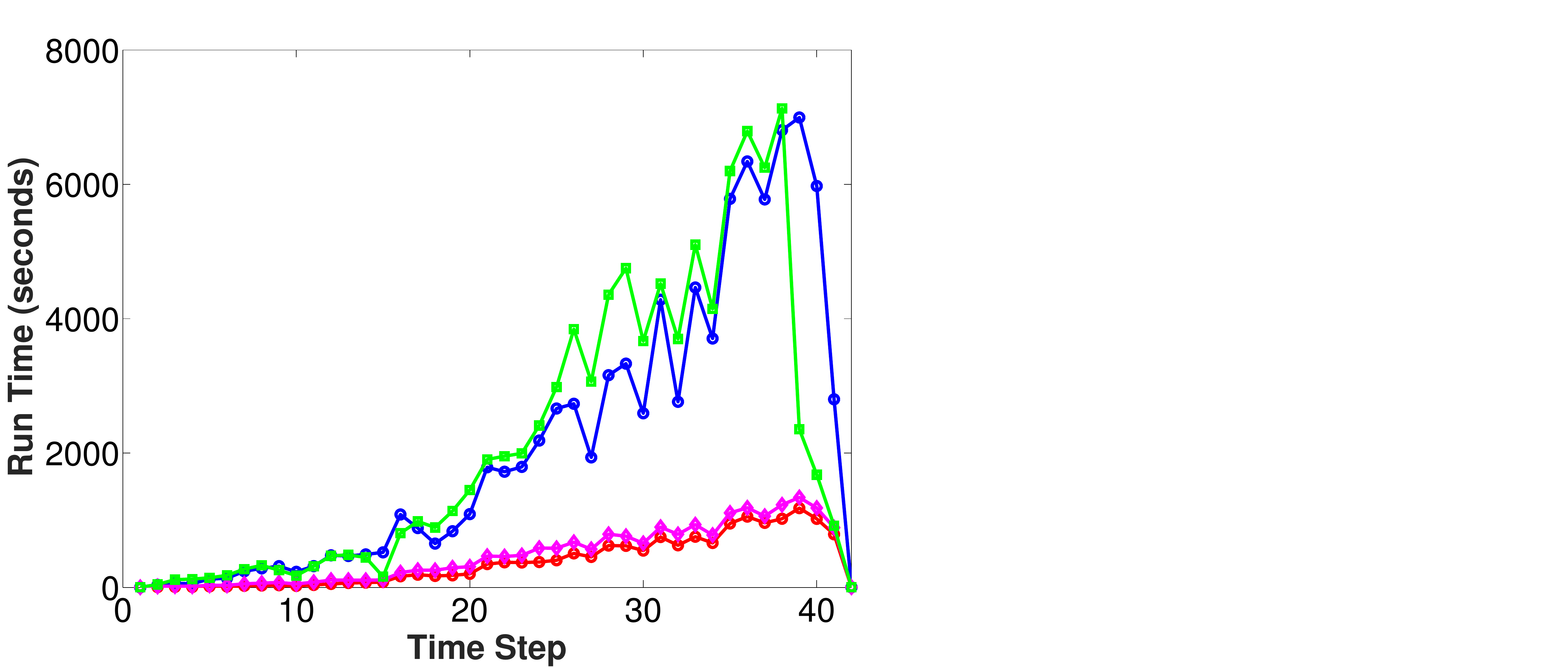}\\
{\small (b) Run Time at every time step}
\end{minipage}
\end{tabular}
\caption{\label{fig:ML1M_rmse_mast}Investigating the merits of side information for MovieLens 1M dataset in the multi-aspect streaming setting. Side information along the user mode is the most useful for tensor completion. See \refsec{sec:ablation} for more details.}
\end{figure}

\begin{figure}
\centering
\begin{tabular}{cc}
\noindent \begin{minipage}[b]{0.5\hsize}
\centering
\includegraphics[width=0.99\textwidth]{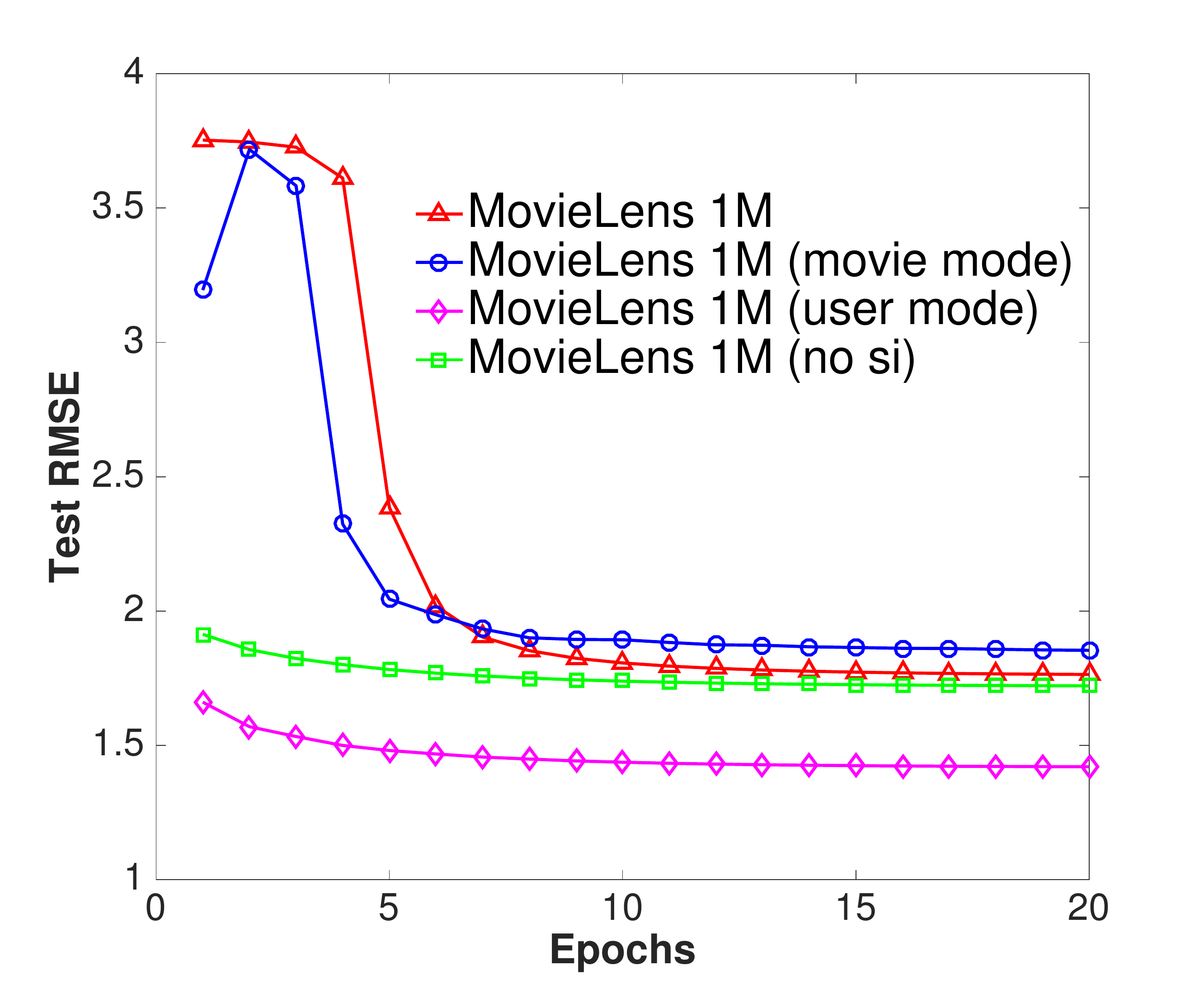}\\
{\small (a) Evolution of Test RMSE against  epochs.}
\end{minipage}
\begin{minipage}[b]{0.5\hsize}
\centering
\includegraphics[width=0.99\textwidth]{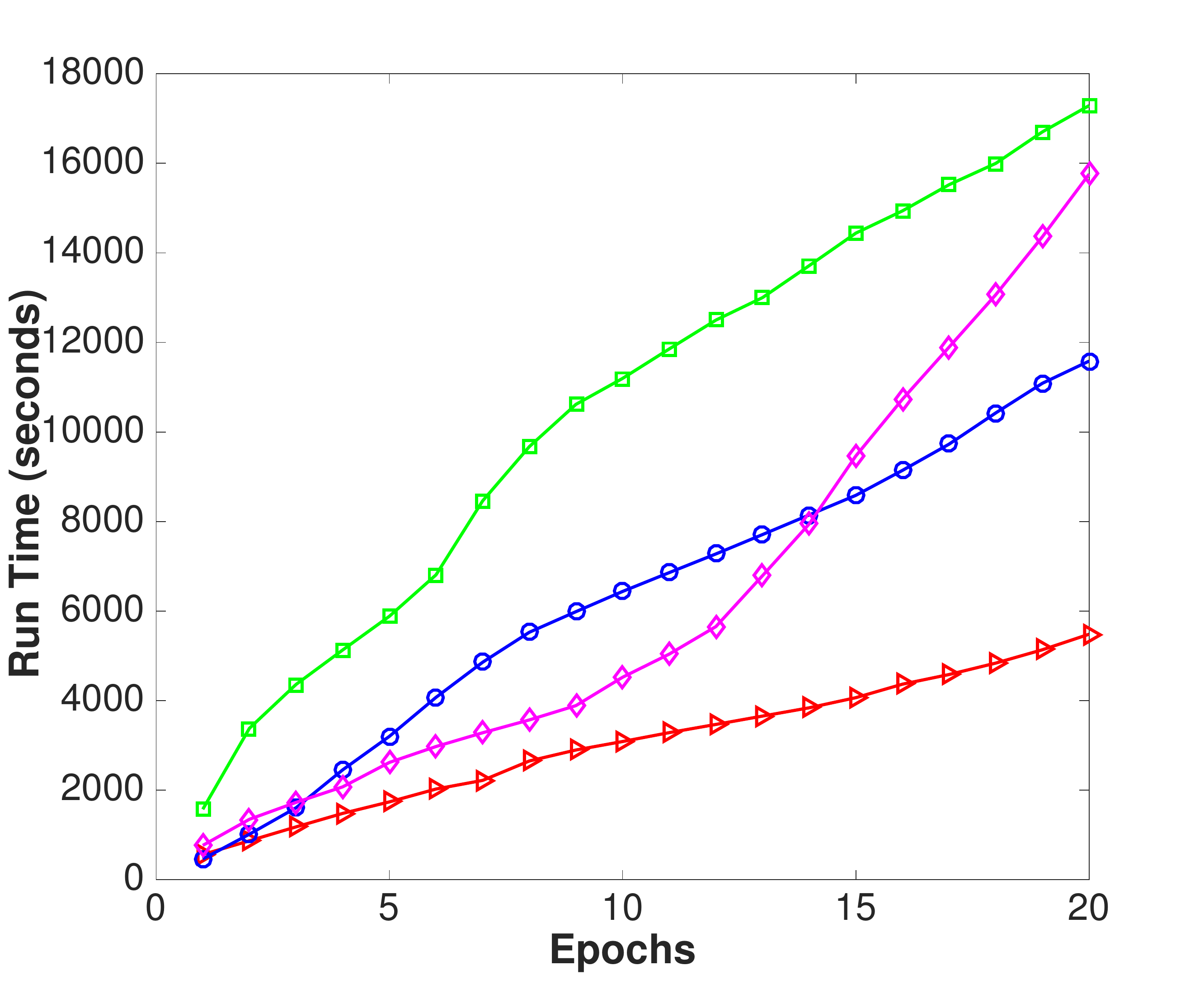}\\
{\small (b) Time elapsed with every epoch.}
\end{minipage}
\end{tabular}
\caption{\label{fig:ablation}Investigating the merits of side information for MovieLens 1M dataset in the batch setting. Side information along the user mode is the most useful for tensor completion. See \refsec{sec:ablation} for more details.}
\end{figure}

Our goal in this paper is to propose a flexible framework using which side information may be easily incorporated during incremental tensor completion, especially in the multi-aspect streaming setting. Our proposed method, \system{}, is motivated by this need. In order to evaluate merits of different types of side information on \system{}, we report several experiments where performances of \system{} with and without various types of side information are compared.

\textbf{Single Side Information}: In the first experiment, we compare \system{} with and without side information (by setting side information to identity; see \refsec{sec:prelim}). We run the experiments in both multi-aspect streaming and streaming settings.
\reftbl{tbl:mast_no_si} reports the mean test RMSE of \system{} and \system{} (w/o SI), which stands for running \system{} without side information, for both datasets in multi-aspect streaming setting. For MovieLens 100K, \system{} achieves better performance without side information. Whereas for YELP, \system{} performs better with side information.
\reffig{fig:rmse_no_si_mast} shows the evolution of test RMSE at every time step. \reffig{fig:runtime_no_si_mast} shows the runtime of \system{} when run with and without side information. 
\system{} runs faster in the presence of side information. \reftbl{tbl:streaming_no_si} reports the mean test RMSE for both the datasets in the streaming setting. Similar to the multi-aspect streaming setting, \system{} achieves better performance without side information for MovieLens 100K dataset and with side information for YELP dataset.
\reffig{fig:rmse_no_si_streaming} shows the test RMSE of \system{} against time steps, with and without side information. \reffig{fig:runtime_no_si_streaming} shows the runtime at every time step.

\textbf{Multi Side Information}: In all the datasets and experiments considered so far, side information along only one mode is available to \system{}. In this next experiment, we consider the setting where side information along multiple modes are available. For this experiment, we consider the {\bf MovieLens 1M } \cite{Harper2015} dataset, a standard  dataset of 1 million movie ratings. This dataset consists of a 6040 (user) $\times$ 3952 (movie)  $\times$ 149 (week) tensor, along with two side information matrices: a 6040 (user) $\times$ 21 (occupation) matrix, and a 3952 (movie) $\times$ 18 (genre) matrix.

Note that among all the methods considered in the paper, \system{} is the only method which scales to the size of MovieLens 1M datasets.

We create four variants of the dataset. The first one with the tensor and all the side information matrices denoted by {MovieLens 1M}, the second one with the tensor and only the side information along the movie mode denoted by {MovieLens 1M (movie mode)}.
Similarly, {MovieLens (user mode)} with only user mode side information, and finally {MovieLens 1M (no si)} with only the tensor and no side information. 

We run \system{} in multi-aspect streaming and batch modes for all the four variants.  Test RMSE at every time step in the multi-aspect streaming setting is shown in \reffig{fig:ML1M_rmse_mast}(a).
Evolution of Test RMSE (lower is better) against epochs are shown in \reffig{fig:ablation}(a) in batch mode. From  Figures \ref{fig:ML1M_rmse_mast}(a) and \ref{fig:ablation}(a), it is evident that the variant {MovieLens 1M (user mode)} achieves best overall performance, implying that the side information along the user mode is more useful for tensor completion in this dataset. However, {MovieLens 1M (movie mode)} achieves poorer performance than other variants implying that movie-mode side information is not useful for tensor completion in this case. This is also the only side information mode available to \system{} during the MovieLens 100K experiments in Tables \ref{tbl:mast_no_si} and \ref{tbl:streaming_no_si}. This sub-optimal side information may be a reason for \system{}'s diminished performance when using side information for MovieLens100K dataset. From the runtime comparisons in Figures \ref{fig:ablation} (b) and \ref{fig:ML1M_rmse_mast}(b), we observe that {MovieLens 1M} (where both types of side information are available) takes the least time, while the variant {MovieLens 1M (no si)} takes the most time to run. This is a benefit we derive from the inductive framework, where in the presence of useful side information, \system{} not only helps in achieving better performance but also runs faster.

\subsection{Unsupervised Setting}
\label{sec:nnsiita_exp}
\begin{figure}
\centering
\begin{tabular}{cc}
\noindent \begin{minipage}[b]{0.5\hsize}
\centering
\includegraphics[scale=0.25]{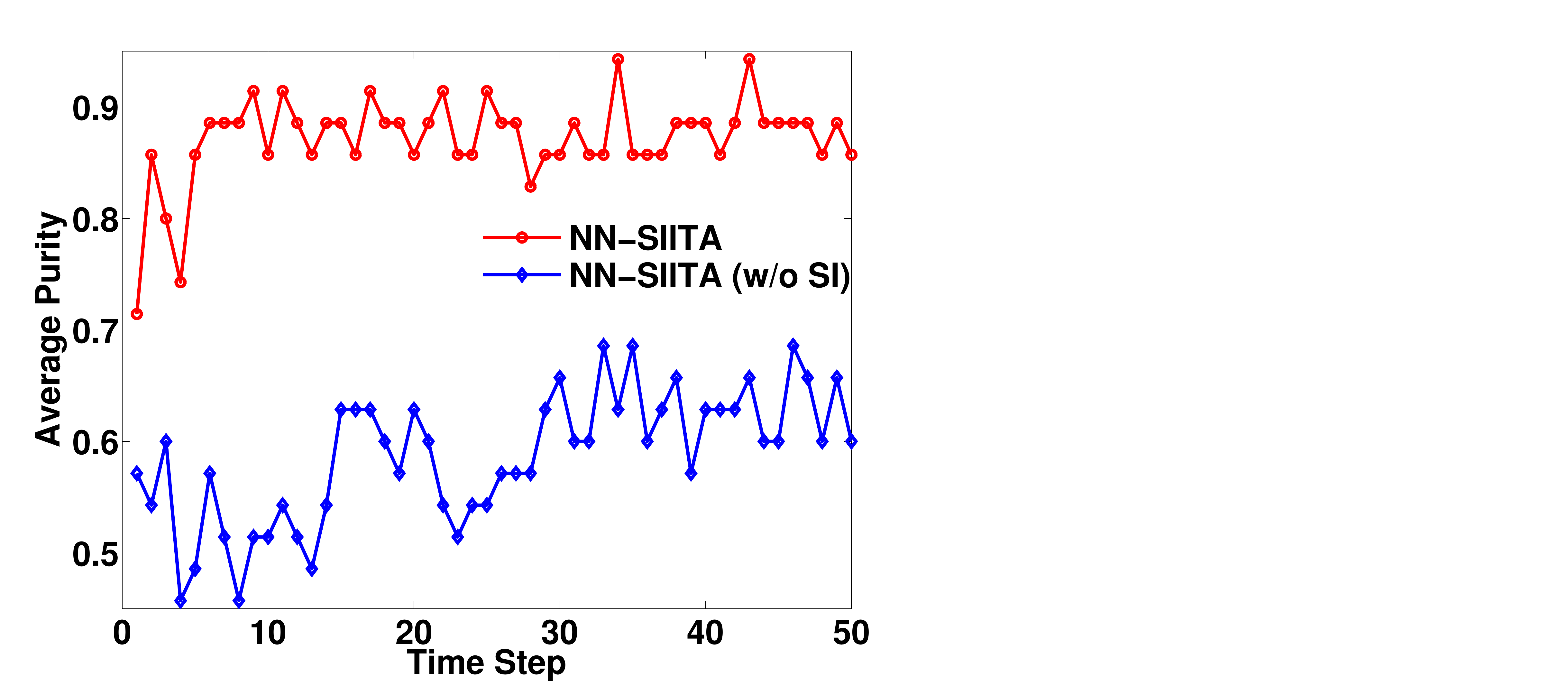}\\
{\small (a) MovieLens 100K}
\end{minipage}
\begin{minipage}[b]{0.5\hsize}
\centering
\includegraphics[scale=0.25]{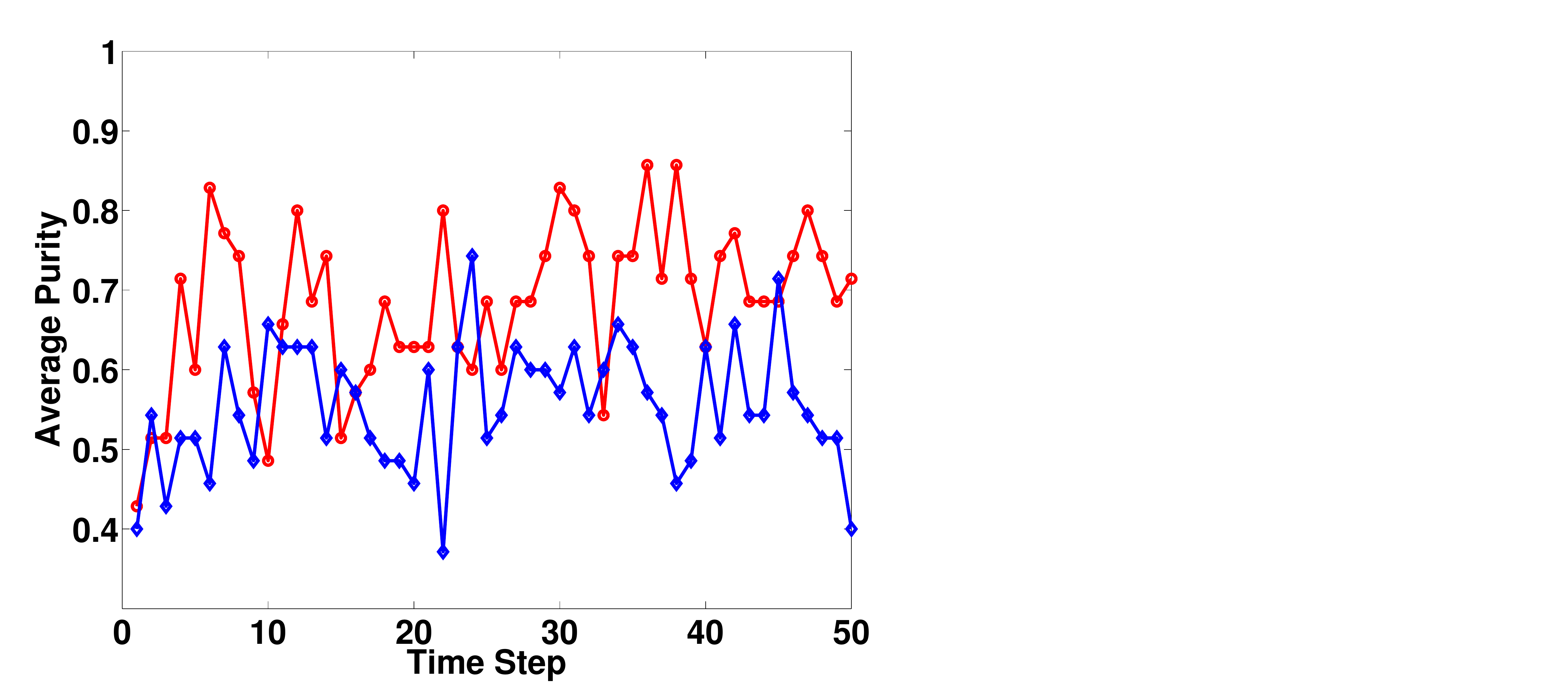}\\
{\small (b) YELP}
\end{minipage}
\end{tabular}
\caption{\label{fig:nn_siita} Average Purity of clusters learned by NN-\system{} and NN-\system{} (w/o SI)  at every time step in the unsupervised setting. For both datasets, side information helps in learning purer clusters. See \refsec{sec:nnsiita_exp} for more details.}
\end{figure}

\begin{figure}
\centering
\begin{tabular}{cc}
\noindent \begin{minipage}[b]{0.5\hsize}
\centering
\includegraphics[scale=0.25]{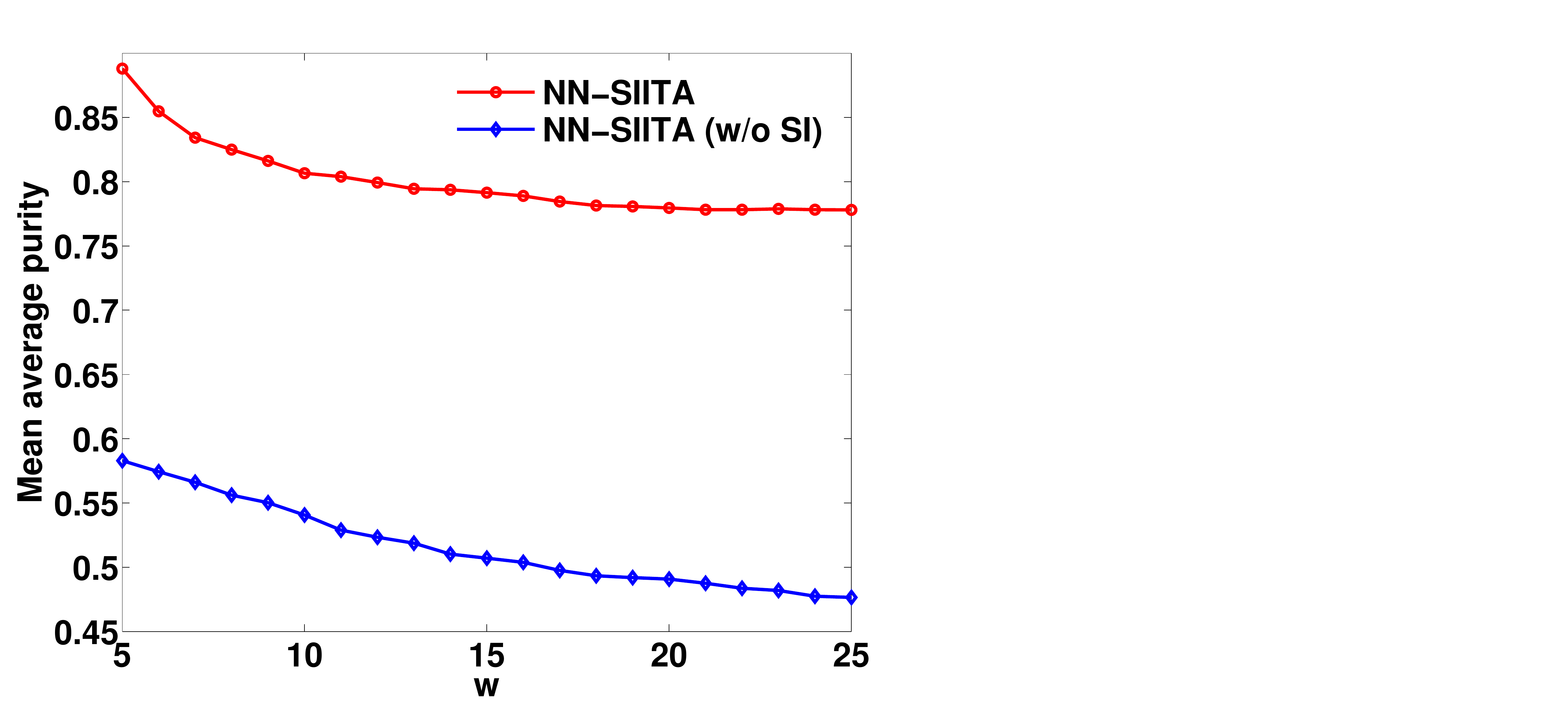}\\
{\small (a) MovieLens 100K}
\end{minipage}
\begin{minipage}[b]{0.5\hsize}
\centering
\includegraphics[scale=0.25]{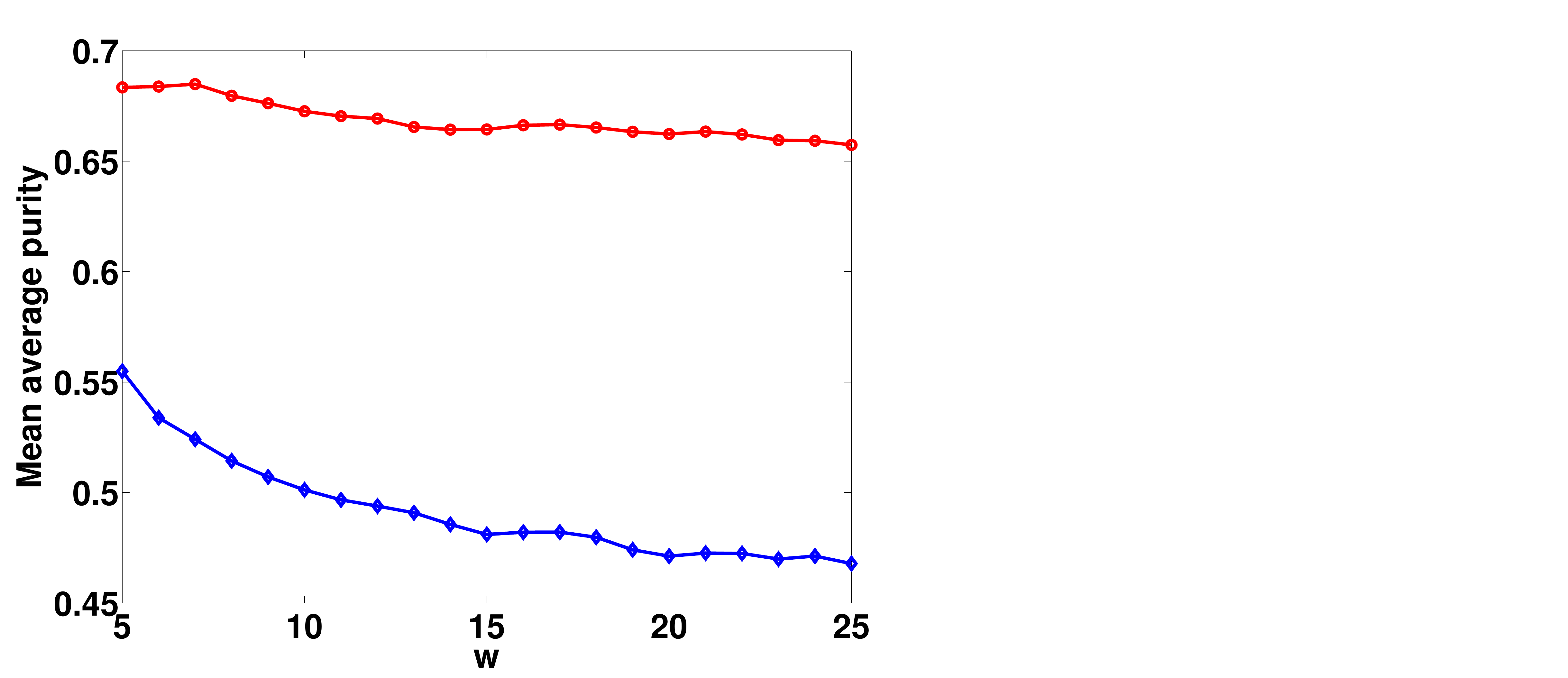}\\
{\small (b) YELP}
\end{minipage}
\end{tabular}
\caption{\label{fig:nn_siita_w} Evolution of mean average purity with $w$ for NN-\system{} and NN-\system{} (w/o SI) for both MovieLens 100K and YELP datasets. See \refsec{sec:nnsiita_exp} for more details.}
\end{figure}

\begin{table*}[tb]
\tiny
\centering
\caption{\label{tbl:nnsita_clusters} Example clusters learned by NN-\system{} for MovieLens 100K and YELP datasets. The first column is an example of a pure cluster and the second column is an example of noisy cluster. See \refsec{sec:nnsiita_exp} for more details.}
 \begin{tabular}{ccccc}
 \toprule
   & \multicolumn{2}{c}{Cluster (Action, Adventure, Sci-Fi)} & \multicolumn{2}{c}{Cluster (Noisy)} \\
  \cline{2-5}
  \multirow{6}{30pt}{MovieLens100K} & \multicolumn{1}{c}{Movie} & \multicolumn{1}{c}{Genres} & \multicolumn{1}{c}{Movie} & \multicolumn{1}{c}{Genres} \\
  \cline{2-5}
				    & The Empire Strikes Back (1980) &   Action, Adventure, Sci-Fi, Drama, Romance  & Toy Story (1995)& Animation, Children's, Comedy\\
				    & Heavy Metal (1981)     &  Action, Adventure, Sci-Fi, Animation, Horror  & From Dusk Till Dawn (1996)& Action, Comedy, Crime, Horror, Thriller\\
				    & Star Wars (1977) & Action, Adventure, Sci-Fi, Romance, War & Mighty Aphrodite (1995)& Comedy\\
				    & Return of the Jedi (1983) & Action, Adventure, Sci-Fi, Romance, War & Apollo 13 (1995)& Action, Drama, Thriller\\
				    & Men in Black (1997) & Action, Adventure, Sci-Fi, Comedy & Crimson Tide (1995)& Drama, Thriller, War \\

\bottomrule
& \multicolumn{2}{c}{Cluster (Phoenix)} & \multicolumn{2}{c}{Cluster (Noisy)} \\
  \cline{2-5}
\multirow{6}{30pt}{YELP} & \multicolumn{1}{c}{Business} & \multicolumn{1}{c}{Location} & \multicolumn{1}{c}{Business} & \multicolumn{1}{c}{Location} \\
\cline{2-5}
			 & Hana Japanese Eatery & \multicolumn{1}{c}{Phoenix} & The Wigman& \multicolumn{1}{c}{Litchfield Park }\\
			 & Herberger Theater Center & \multicolumn{1}{c}{Phoenix} & Hitching Post 2 &  \multicolumn{1}{c}{Gold Canyon }\\
			 & Scramble A Breakfast Joint & \multicolumn{1}{c}{Phoenix}& Freddys Frozen Custard \& Steakburgers& \multicolumn{1}{c}{Glendale}\\
			 & The Arrogant Butcher  & \multicolumn{1}{c}{Phoenix}& Costco& \multicolumn{1}{c}{Avondale}\\
			 & FEZ & \multicolumn{1}{c}{Phoenix}& Hana Japanese Eatery& \multicolumn{1}{c}{Phoenix}\\
\bottomrule
 \end{tabular}
\end{table*}

In this section, we consider an unsupervised setting with the aim to discover underlying clusters of the items, like movies in the MovieLens 100K dataset and businesses in the YELP dataset, from a sequence of sparse tensors. It is desirable to mine clusters such that similar items are grouped together. Nonnegative constraints are essential for mining interpretable clusters \cite{Hyvoenen2008, Murphy2012}. For this set of experiments, we consider the nonnegative version of \system{} denoted by NN-\system{}. We investigate whether side information helps in discovering more coherent clusters of items in both datasets.

We run our experiments in the multi-aspect streaming setting. At every time step, we compute {\it Purity} of clusters and report average-Purity. Purity of a cluster is defined as the percentage of the cluster that is coherent. For example, in MovieLens 100K, a cluster of movies is 100\% pure if all the movies belong to the same genre and 50\% pure if only half of the cluster belong to the same genre. Formally, let clusters of items along mode-$i$ are desired, let $r_i$ be the rank of factorization along mode-$i$. Every column of the matrix $\mat{A}_i \mat{U}_i$ is considered a distribution of the items, the top-$w$ items of the distribution represent a cluster. For $p$-th cluster, i.e., cluster representing column $p$ of the matrix  $\mat{A}_i \mat{U}_i$, let $w_p$ items among the top-$w$ items belong to the same category, Purity and average-Purity are defined as follows:

\begin{align*}
 \text{Purity}(p) = w_p/w, \\
 \text{average-Purity} = \frac{1}{r_i} \sum\limits_{p=1}^{r_i} \text{Purity}(p).
\end{align*}


Note that Purity is computed per cluster, while average-Purity is computed for a set of clusters. Higher average-Purity indicates a better clustering. 

We report average-Purity at every time step for both the datasets. We run NN-\system{} with and without side information. \reffig{fig:nn_siita} shows average-Purity at every time step for MovieLens 100K and YELP datasets. It is clear from \reffig{fig:nn_siita} that for both the datasets side information helps in discovering better clusters. We compute the Purity for MovieLens 100K dataset based on the genre information of the movies and for the YELP dataset we compute Purity based on the geographic locations of the businesses. \reftbl{tbl:nnsita_clusters} shows some example clusters learned by NN-\system{}. For MovieLens 100K dataset, each movie can belong to multiple genres. For computing the Purity, we consider the most common genre for all the movies in a cluster. Results shown in \reffig{fig:nn_siita} are for $w = 5$. However, we also vary $w$ between 5 and 25 and report the \emph{mean} average-Purity, which is obtained by computing the mean across all the time steps in the multi-aspect streaming setting.  As can be seen from \reffig{fig:nn_siita_w}, having side information helps in learning better clusters for all the values of $w$. For MovieLens 100K, the results reported are with a factorization rank of $(3,7,3)$ and for YELP, the rank of factorization is $(5,7,3)$. Since this is an unsupervised setting, note that we use the entire data for factorization, i.e., there is no train-test split.

\section{Conclusion}
\label{sec:conclusion}
We propose an inductive framework for incorporating side information for tensor completion in standard and multi-aspect streaming settings. The proposed framework can also be used in the batch setting. Given a completely new dataset with side information along multiple modes, \system{} can be used to analyze the merits of different side information for tensor completion. Besides performing better, \system{} is also significantly faster than state-of-the-art algorithms. We also propose NN-\system{} for handling nonnegative constraints and show how it can be used for mining interpretable clusters. Our experiments confirm the effectiveness of \system{} in many instances. In future, we plan to extend our proposed framework to handle missing side information problem instances \cite{Kishan2017}.

\subsection*{ Acknowledgement } This work is supported in part by the Ministry of Human Resource Development (MHRD), Government of India.

\bibliographystyle{authordate1}
\bibliography{arXiv_si_ita}

\begin{thebibliography}{}

\bibitem[\protect\citename{Acar {\em et~al.\ }\relax, }2011]{Acar2011}
Acar, Evrim, Kolda, Tamara~G., \& Dunlavy, Daniel~M. 2011.
\newblock All-at-once Optimization for Coupled Matrix and Tensor
  Factorizations.
\newblock {\em In:} {\em MLG}.

\bibitem[\protect\citename{Beutel {\em et~al.\ }\relax, }2014]{Beutel2014}
Beutel, Alex, Talukdar, Partha~Pratim, Kumar, Abhimanu, Faloutsos, Christos,
  Papalexakis, Evangelos~E, \& Xing, Eric~P. 2014.
\newblock Flexifact: Scalable flexible factorization of coupled tensors on
  hadoop.
\newblock {\em In:} {\em SDM}.

\bibitem[\protect\citename{Cichocki {\em et~al.\ }\relax, }2015]{Cichocki2015}
Cichocki, A., Mandic, D., De~Lathauwer, L., Zhou, G., Zhao, Q., Caiafa, C., \&
  Phan, H.~A. 2015.
\newblock Tensor decompositions for signal processing applications: From
  two-way to multiway component analysis.
\newblock {\em IEEE Signal Processing Magazine}, {\bf 32}(2), 145--163.

\bibitem[\protect\citename{Ermi{\c{s}} {\em et~al.\ }\relax, }2015a]{Ermis2015}
Ermi{\c{s}}, B., Acar, E., \& Cemgil, A.~T. 2015a.
\newblock Link prediction in heterogeneous data via generalized coupled tensor
  factorization.
\newblock {\em In:} {\em KDD}.

\bibitem[\protect\citename{Ermi{\c{s}} {\em et~al.\ }\relax,
  }2015b]{Ermis2015a}
Ermi{\c{s}}, Beyza, Acar, Evrim, \& Cemgil, A~Taylan. 2015b.
\newblock Link prediction in heterogeneous data via generalized coupled tensor
  factorization.
\newblock {\em KDD}.

\bibitem[\protect\citename{Fanaee-T \& Gama, }2015]{Fanaee-T2015}
Fanaee-T, Hadi, \& Gama, Jo\~{a}o. 2015.
\newblock Multi-aspect-streaming Tensor Analysis.
\newblock {\em Know.-Based Syst.},  332--345.

\bibitem[\protect\citename{Filipovi{\'{c}} \& Juki{\'{c}},
  }2015]{Filipovic2015}
Filipovi{\'{c}}, M., \& Juki{\'{c}}, A. 2015.
\newblock Tucker factorization with missing data with application to low-n-rank
  tensor completion.
\newblock {\em Multidimens Syst Signal Process}.

\bibitem[\protect\citename{Ge {\em et~al.\ }\relax, }2016]{Ge2016}
Ge, Hancheng, Caverlee, James, Zhang, Nan, \& Squicciarini, Anna. 2016.
\newblock Uncovering the Spatio-Temporal Dynamics of Memes in the Presence of
  Incomplete Information.
\newblock CIKM.

\bibitem[\protect\citename{Guo {\em et~al.\ }\relax, }2017]{Guo2017}
Guo, X., Yao, Q., \& Kwok, J.~T. 2017.
\newblock Efficient sparse low-rank tensor completion using the Frank-Wolfe
  algorithm.
\newblock {\em In:} {\em AAAI}.

\bibitem[\protect\citename{Harper \& Konstan, }2015]{Harper2015}
Harper, F.~Maxwell, \& Konstan, Joseph~A. 2015.
\newblock The MovieLens Datasets: History and Context.
\newblock {\em ACM Trans. Interact. Intell. Syst.}, Dec., 19:1--19:19.

\bibitem[\protect\citename{Hyv{\"o}nen {\em et~al.\ }\relax,
  }2008]{Hyvoenen2008}
Hyv{\"o}nen, Saara, Miettinen, Pauli, \& Terzi, Evimaria. 2008.
\newblock Interpretable nonnegative matrix decompositions.
\newblock {\em Pages  345--353 of:} {\em KDD}.
\newblock ACM.

\bibitem[\protect\citename{Jain \& Dhillon, }2013]{jain2013}
Jain, Prateek, \& Dhillon, Inderjit~S. 2013.
\newblock Provable inductive matrix completion.
\newblock {\em arXiv preprint arXiv:1306.0626}.

\bibitem[\protect\citename{Jeon {\em et~al.\ }\relax, }2016]{Jeon2016}
Jeon, ByungSoo, Jeon, Inah, Sael, Lee, \& Kang, U. 2016.
\newblock Scout: Scalable coupled matrix-tensor factorization-algorithm and
  discoveries.
\newblock {\em In:} {\em ICDE}.

\bibitem[\protect\citename{Kasai \& Mishra, }2016]{Kasai2016a}
Kasai, H., \& Mishra, B. 2016.
\newblock Low-rank tensor completion: a Riemannian manifold preconditioning
  approach.
\newblock {\em In:} {\em ICML}.

\bibitem[\protect\citename{Kasai, }2016]{Kasai2016}
Kasai, Hiroyuki. 2016.
\newblock Online Low-Rank Tensor Subspace Tracking from Incomplete Data by CP
  Decomposition using Recursive Least Squares.
\newblock {\em In:} {\em ICASSP}.

\bibitem[\protect\citename{Kim \& Choi, }2007]{Kim2007}
Kim, Yong-Deok, \& Choi, Seungjin. 2007.
\newblock Nonnegative tucker decomposition.
\newblock {\em In:} {\em CVPR}.

\bibitem[\protect\citename{Kolda \& Bader, }2009]{Kolda2009}
Kolda, Tamara~G, \& Bader, Brett~W. 2009.
\newblock Tensor decompositions and applications.
\newblock {\em SIAM review}, {\bf 51}(3), 455--500.

\bibitem[\protect\citename{Lefevre {\em et~al.\ }\relax, }2011]{Lefevre2011}
Lefevre, Augustin, Bach, Francis, \& F{\'e}votte, C{\'e}dric. 2011.
\newblock Online algorithms for nonnegative matrix factorization with the
  Itakura-Saito divergence.
\newblock {\em Pages  313--316 of:} {\em Applications of Signal Processing to
  Audio and Acoustics (WASPAA), 2011 IEEE Workshop on}.
\newblock IEEE.

\bibitem[\protect\citename{Mardani {\em et~al.\ }\relax, }2015]{Mardani2015}
Mardani, Morteza, Mateos, Gonzalo, \& Giannakis, Georgios~B. 2015.
\newblock Subspace learning and imputation for streaming big data matrices and
  tensors.
\newblock {\em IEEE Transactions on Signal Processing}.

\bibitem[\protect\citename{M{\o}rup {\em et~al.\ }\relax, }2008]{Morup2008}
M{\o}rup, Morten, Hansen, Lars~Kai, \& Arnfred, Sidse~M. 2008.
\newblock Algorithms for sparse nonnegative Tucker decompositions.
\newblock {\em Neural computation}, {\bf 20}(8), 2112--2131.

\bibitem[\protect\citename{Murphy {\em et~al.\ }\relax, }2012]{Murphy2012}
Murphy, Brian, Talukdar, Partha~Pratim, \& Mitchell, Tom~M. 2012.
\newblock Learning Effective and Interpretable Semantic Models using
  Non-Negative Sparse Embedding.
\newblock {\em In:} {\em COLING}.

\bibitem[\protect\citename{Narita {\em et~al.\ }\relax, }2011]{Narita2011}
Narita, Atsuhiro, Hayashi, Kohei, Tomioka, Ryota, \& Kashima, Hisashi. 2011.
\newblock Tensor Factorization Using Auxiliary Information.
\newblock {\em Pages  501--516 of:} {\em Machine Learning and Knowledge
  Discovery in Databases}.

\bibitem[\protect\citename{Natarajan \& Dhillon, }2014]{Natarajan2014}
Natarajan, Nagarajan, \& Dhillon, Inderjit~S. 2014.
\newblock Inductive matrix completion for predicting gene--disease
  associations.
\newblock {\em Bioinformatics}, {\bf 30}(12), i60--i68.

\bibitem[\protect\citename{Nimishakavi {\em et~al.\ }\relax,
  }2016]{Nimishakavi2016}
Nimishakavi, Madhav, Saini, Uday~Singh, \& Talukdar, Partha. 2016.
\newblock Relation Schema Induction using Tensor Factorization with Side
  Information.
\newblock {\em Pages  414--423 of:} {\em EMNLP}.

\bibitem[\protect\citename{Nion \& Sidiropoulos, }2009]{Nion2009}
Nion, Dimitr, \& Sidiropoulos, {Nicholas D.} 2009.
\newblock Adaptive algorithms to track the PARAFAC decomposition of a
  third-order tensor.
\newblock {\em IEEE Transactions on Signal Processing}.

\bibitem[\protect\citename{Shashua \& Hazan, }2005]{Shashua2005}
Shashua, Amnon, \& Hazan, Tamir. 2005.
\newblock Non-negative Tensor Factorization with Applications to Statistics and
  Computer Vision.
\newblock {\em Pages  792--799 of:} {\em ICML}.
\newblock ICML '05.
\newblock New York, NY, USA: ACM.

\bibitem[\protect\citename{Si {\em et~al.\ }\relax, }2016]{Si2016}
Si, Si, Chiang, Kai-Yang, Hsieh, Cho-Jui, Rao, Nikhil, \& Dhillon, Inderjit~S.
  2016.
\newblock Goal-directed inductive matrix completion.
\newblock {\em In:} {\em KDD}.

\bibitem[\protect\citename{Song {\em et~al.\ }\relax, }2017]{Song2017}
Song, Qingquan, Huang, Xiao, Ge, Hancheng, Caverlee, James, \& Hu, Xia. 2017.
\newblock Multi-Aspect Streaming Tensor Completion.
\newblock {\em In:} {\em KDD}.

\bibitem[\protect\citename{Sun {\em et~al.\ }\relax, }2006]{Sun2006}
Sun, Jimeng, Tao, Dacheng, \& Faloutsos, Christos. 2006.
\newblock Beyond streams and graphs: dynamic tensor analysis.
\newblock {\em In:} {\em KDD}.

\bibitem[\protect\citename{Sun {\em et~al.\ }\relax, }2008]{Sun2008}
Sun, Jimeng, Tao, Dacheng, Papadimitriou, Spiros, Yu, Philip~S., \& Faloutsos,
  Christos. 2008.
\newblock Incremental Tensor Analysis: Theory and Applications.
\newblock {\em ACM Trans. Knowl. Discov. Data}, {\bf 2}(3).

\bibitem[\protect\citename{Symeonidis {\em et~al.\ }\relax,
  }2008]{Symeonidis2008}
Symeonidis, Panagiotis, Nanopoulos, Alexandros, \& Manolopoulos, Yannis. 2008.
\newblock Tag Recommendations Based on Tensor Dimensionality Reduction.
\newblock {\em In:} {\em RecSys}.

\bibitem[\protect\citename{Welling \& Weber, }2001]{Welling2001}
Welling, Max, \& Weber, Markus. 2001.
\newblock Positive tensor factorization.
\newblock {\em Pattern Recognition Letters}, {\bf 22}(12), 1255--1261.

\bibitem[\protect\citename{Wimalawarne {\em et~al.\ }\relax, }2017]{Kishan2017}
Wimalawarne, Kishan, Yamada, Makoto, \& Mamitsuka, Hiroshi. 2017.
\newblock Convex Coupled Matrix and Tensor Completion.
\newblock {\em arXiv preprint arXiv:1705.05197}.

\bibitem[\protect\citename{Yu {\em et~al.\ }\relax, }2015]{Yu2015}
Yu, Rose, Cheng, Dehua, \& Liu, Yan. 2015.
\newblock Accelerated Online Low-rank Tensor Learning for Multivariate
  Spatio-temporal Streams.
\newblock {\em In:} {\em ICML}.

\bibitem[\protect\citename{Zhao {\em et~al.\ }\relax, }2017]{Zhao2017}
Zhao, Renbo, Tan, Vincent, \& Xu, Huan. 2017.
\newblock Online Nonnegative Matrix Factorization with General Divergences.
\newblock {\em Pages  37--45 of:} {\em AISTATS}.

\bibitem[\protect\citename{Zhou {\em et~al.\ }\relax, }2016]{Zhou2016}
Zhou, Shuo, Vinh, Nguyen~Xuan, Bailey, James, Jia, Yunzhe, \& Davidson, Ian.
  2016.
\newblock Accelerating online cp decompositions for higher order tensors.
\newblock {\em In:} {\em KDD}.
\newblock ACM.

\end{thebibliography}

\end{document}